\documentclass[pdflatex,sn-basic,iicol]{sn-jnl}% Vancouver Numbered Reference Style

\usepackage{graphicx}%
\usepackage{multirow}%
\usepackage{amsmath,amssymb,amsfonts}%
\usepackage{mathtools}% for \text{argmin}
\usepackage{amsthm}%
\usepackage{mathrsfs}%
\usepackage[title]{appendix}%
\usepackage{xcolor}%
\usepackage{textcomp}%
\usepackage{manyfoot}%
\usepackage{booktabs}%
\usepackage{algorithm}%
\usepackage{algorithmicx}%
\usepackage{algpseudocode}%
\usepackage{listings}%
\usepackage{multirow}
\usepackage{bm}
\usepackage{amsmath}

\usepackage{tabularx}
\usepackage{hyperref}       % hyperlinks
\usepackage{url}            % simple URL typesetting     % blackboard math symbols
\usepackage{nicefrac}       % compact symbols for 1/2, etc.
\usepackage[expansion=false]{microtype}
\usepackage{lipsum}
\usepackage{fancyhdr}       % header
\usepackage{comment}
\usepackage{placeins}
\usepackage{multicol}
\usepackage{caption}
\usepackage{newtxtext,newtxmath}

\theoremstyle{thmstyleone}%
\theoremstyle{thmstyletwo}%

\theoremstyle{thmstylethree}%

\begin{document}

\title[Materialist]{Materialist: Physically Based Editing Using Single-Image Inverse Rendering}

\author*[1]{\fnm{Lezhong} \sur{Wang}}\email{lewa@dtu.dk}
\author[1]{\fnm{Duc Minh} \sur{Tran}}\email{dmitr@dtu.dk}
\author[1]{\fnm{Ruiqi} \sur{Cui}}\email{ruicu@dtu.dk}
\author[1]{\fnm{Thomson} \sur{TG}}\email{thtg@dtu.dk}
\author[1]{\fnm{Anders Bjorholm} \sur{Dahl}}\email{abda@dtu.dk}
\author[1]{\fnm{Siavash Arjomand} \sur{Bigdeli}}\email{sarbi@dtu.dk}
\author[1]{\fnm{Jeppe Revall} \sur{Frisvad}}\email{jerf@dtu.dk}
\author[2]{\fnm{Manmohan} \sur{Chandraker}}\email{mkchandraker@ucsd.edu}

\affil*[1]{\orgname{Technical University of Denmark}, \orgaddress{\city{Kongens Lyngby}, \country{Denmark}}}

\affil[2]{\orgname{University of California San Diego}, \orgaddress{\city{San Diego}, \country{United States of America}}}

\abstract{Achieving physically consistent image editing remains a significant challenge in computer vision.
Existing image editing methods typically rely on neural networks, which struggle to accurately handle shadows and refractions. Conversely, physics-based inverse rendering often requires multi-view optimization, limiting its practicality in single-image scenarios.
In this paper, we propose \textbf{Materialist}, a neural-initialized physically based rendering pipeline for single-image inverse rendering. Unlike previous hybrid methods that use physics to guide neural generation, our method leverages neural networks to predict initial material properties, which are then rigorously optimized via progressive differentiable rendering.
Our approach enables a range of applications, including material editing, object insertion, and relighting, while also introducing an effective method for editing material transparency via ray-traced refraction without requiring full scene geometry. Furthermore, our envmap estimation method also achieves competitive performance, further enhancing the accuracy of image editing task.
Experiments demonstrate strong performance across synthetic and real-world datasets, excelling even on challenging out-of-domain images.}


\maketitle

\begin{figure*}
    \centering
    \includegraphics[width=1.0\textwidth]{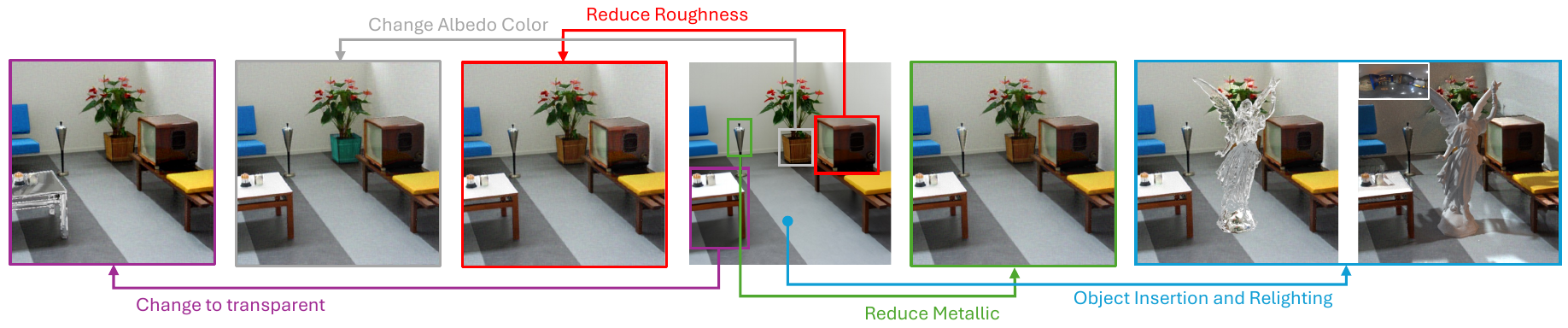} 
    \captionof{figure}{Given an image (in the middle), our inverse rendering approach enables physically based image editing, including transparency, albedo, roughness, metallic, transparent object insertion and relighting. 
    The image is sourced from the IIW dataset \citep{bell14intrinsic}. %image No. 34251.
    }
    \label{fig:teaser}
\end{figure*}

\section{Introduction} \label{sec:intro}

\begin{figure}[tb]
    \centering
    \includegraphics[width=\linewidth]{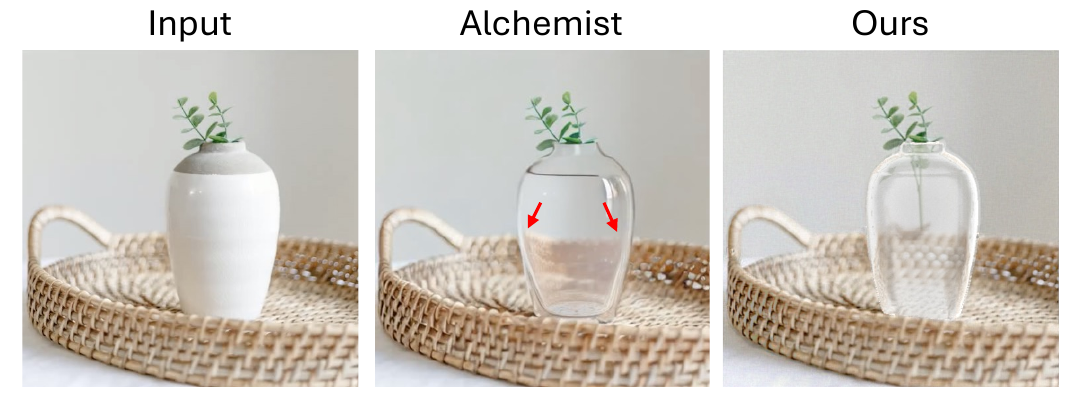}
    \caption{An image from the project website of Alchemist \citep{sharma2024alchemist}. This is an example of a refraction failure case.
    While the diffusion-based method generates high-frequency specular highlights, it fails to respect the scene's physical reality: note the disappearance of the internal plant and the incorrect refraction of the background. Our method physically simulates light transport through the object geometry. Although the lighting estimation is challenging in single-view scenarios, our approach correctly preserves the internal content (the plant) and renders physically consistent background refraction.  
    }
    \label{fig:refraction_failure}
\end{figure}

High-quality image editing requires professional skills. Reducing the complexity and increasing the accuracy of image editing has long been a focus in computer vision and computer graphics~\citep{popat1993novel,barnes2009patchmatch,mengsdedit,pan2023drag}. With the success of the generative Diffusion Model (DM) \citep{rombach2022high} in image generation, researchers have explored use of DM for image editing~\citep{hertzprompt,mokady2023null,brooks2023instructpix2pix}. However, a DM often struggles with precise material property editing. 
Alchemist~\citep{sharma2024alchemist} attempts to address this by training DM on a large synthetic dataset. 
However, while Alchemist operates on physically based material maps, its inference relies on a probabilistic diffusion process that does not explicitly enforce physical laws during generation, such as ray optics or strict shadow consistency. 
Consequently, it often struggles with complex light transport effects, such as generating accurate refractive distortions for transparent objects (as demonstrated in Fig.~\ref{fig:refraction_failure}).

Inverse rendering offers a promising solution to these challenges. Single-view inverse rendering~\citep{sengupta2019neural,li2021openrooms,li2022physically,zhu2022irisformer} decomposes images into albedo, roughness, and metallic properties via neural networks (MaterialNet), enhancing interpretability. However, relying on a neural renderer as a substitute for a physically based renderer still shares limitations with DM, such as difficulties in accurately handling of shadows and refraction. Multi-view inverse rendering~\citep{azinovic2019inverse,yao2022neilf,wu2023factorized,yu2023milo,li2023multi,bao2024photometric} leverages multi-view constraints and the rendering equation to adhere better to physical principles and achieve superior results. The multi-view input requirement, however, restricts its applicability. Applying differentiable rendering to a single image leads to an ill-posed problem with infinitely many possible solutions.

To address these issues, we adopt a more interpretable inverse rendering approach that combines the strengths of learning and physically based rendering. Additionally, we introduce material transparency editing, a capability not previously explored in single-image inverse rendering.

Our main contributions are:
\begin{itemize}
     \item \textbf{Neural-Initialized Physical Optimization}: We propose a novel inverse rendering pipeline that utilizes neural priors to initialize a rigorous differentiable ray-tracing optimization. Unlike retrieval-based methods~\citep{yan2023psdr} or physics-guided neural generation~\citep{careaga2025physically}, our approach performs in-place reconstruction and ensures the final output is generated by a physics-based engine, guaranteeing physical consistency.

    \item \textbf{Better Envmap Estimation}: Benefiting from MaterialNet's predictions of material properties, we achieve better results in single-view differentiable Monte Carlo ray tracing for environment maps.

    \item \textbf{Explicit Physically Based Editing}: Instead of relying on a neural renderer which may generate unphysical effects, our method leverages a physically based renderer~\citep{jakob2022mitsuba3} to achieve strictly physically based material editing. This ensures precise global illumination, shadows, and realistic object-environment interactions, thus achieving competitive performance in material editing, relighting, and transparent object insertion tasks.
    
    \item \textbf{Geometry-Aware Transparency Editing}: We propose a single-view, physically based transparency editing method. By ray-tracing through the reconstructed mesh to model Index of Refraction (IOR) and specular transmission, we achieve realistic refractive distortions that purely 2D image-based methods \citep{khan2006image} cannot capture, all without requiring full scene geometry.

\end{itemize}
These contributions are summarized in relation to existing work in Table~\ref{tab:novelty_comparison}.
In addition, our material prediction model achieve competitive performance compared to latest DM-based model~\citep{kocsis2024intrinsic,zeng2024rgb} on synthetic datasets, running faster than diffusion-based estimators. This efficiency enables a rapid direct inference mode, while also serving as a robust initialization for our progressive differentiable rendering.
Furthermore, to ensure artifact-free rendering at boundaries, we integrate a boundary-aware mesh reconstruction strategy adapted from layered depth representations~\citep{shade1998layered, shih20203d}.

\begin{figure}
    \centering
    \includegraphics[width=0.7\linewidth]{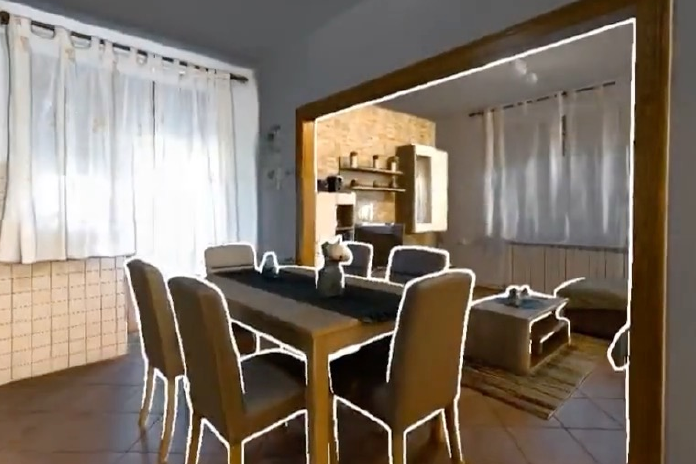}
    \caption{A mesh reconstructed by MoGe \citep{wang2025moge} exhibits a white border at the breakpoints between the foreground and background}
    \label{fig:moge}
\end{figure}

\section{Related Work} \label{sec:related}

\paragraph{Multi-View Inverse Rendering}
Early work like inverse path tracing~\citep{azinovic2019inverse} was limited to simple scenes. They used multi-view constraints and assumed known scene geometry when using the rendering equation for inverse rendering. With NeRF~\citep{mildenhall2021nerf}, physically based inverse rendering with multi-view images became more feasible~\citep{yao2022neilf,yu2023milo,li2023multi,sun2023neural,bao2024photometric}. In addition, use of differentiable Monte Carlo rendering led to better shape and material estimation~\citep{luan2021unified,wang2021neus,hasselgren2022shape,zhang2022iron,yu2022monosdf,wu2023factorized}, but these techniques %FIPT~\citep{wu2023factorized} improved IPT by optimizing material properties while baking lighting, rather than jointly optimizing both. However, FIPT typically need pre-reconstructed scene geometry via NeRF-like methods~\citep{yu2022monosdf,wang2021neus} and
still rely on multi-view inputs. 
More recent methods employ neural networks as renderers for multi-view inputs~\citep{choi2023mair}, bypassing computational optimization but currently lacking the performance of equation-based methods in material editing and object insertion. Like most multi-view techniques, our approach uses the rendering equation but requires only a single-view image as input.

\paragraph{Single-View Inverse Rendering}
Unlike multi-view inverse rendering, which relies on the rendering equation, single-view inverse rendering often uses neural networks trained on large datasets to replace physically based rendering~\citep{li2018learning,li2020inverse,li2022physically,sengupta2019neural,sang2020single,zhu2022irisformer,luo2024intrinsicdiffusion}. A neural network usually called MaterialNet, or MatNet, predicts material properties (albedo, roughness, metallic) and per-pixel lighting from a single image, which are then used for image synthesis. Although neural renderers perform well in specific tasks like human face relighting~\citep{pandey2021total,yeh2022learning}, they struggle with generalist tasks where physical simulation is more accurate such as material editing and transparent object insertion. Unless trained on a well represented dataset, the neural renderers fall apart. 

Unlike optimization-by-retrieval methods like PSDR-Room \citep{yan2023psdr}, which replace image content with database assets, our method optimizes per-pixel material properties to preserve the original object's identity.

In differentiable rendering, early studies addressed rasterization non-differentiability~\citep{chen2019dibrender,chen2021dib}, while recent work has explored inverse Monte Carlo ray tracing~\citep{kato2020differentiable,zhang2021path,deng2022reconstructing}, similar to our physics-based optimization. However, these methods require significant priors and are limited to single objects, making them less suitable for complex scenes.
Our approach also uses MatNet for single-view material property inference but incorporates a physically based renderer.

\begin{table*}[!htb]
\centering
\caption{Comparison of contributions. Our method is the only one that simultaneously supports general object editing, physically based transparency, and explicitly handles mesh boundary artifacts.}
\label{tab:novelty_comparison}
\renewcommand{\arraystretch}{1.3}
\resizebox{\textwidth}{!}{%
\begin{tabular}{l|p{3.5cm}|p{3.5cm}|p{3.5cm}|p{3.5cm}}
\toprule
Feature / Method & \cite{khan2006image} & \cite{yan2023psdr} & \cite{careaga2025physically} & Ours \\ \hline
Target Domain & General (Image-based) & Indoor Scenes (Room Layout) & General Scenes & General Objects \& Scenes \\ \hline
Core Methodology & 2D Image Filters & 3D Retrieval & Physics-Guided Neural Generation & Neural-Initialized PBR \\ \hline
Physics-Based & No  & Yes & Assumes Lambertian surfaces & Disney BSDF \\ \hline
Transparency Editing & 2D Distortion & No & No & Ray-Traced Refraction \\ \hline
Geometry Awareness & None & Room Layout / Planes & Mesh from Intrinsic &  Border-free 3D reconstruction \\ \hline
Primary Goal & Visual Effect & Indoor Scene Reconstruction & Relighting & Material Editing (Inc. Trans.) + Relighting \\ \bottomrule
\end{tabular}
}
\vspace{2ex}
\end{table*}

\paragraph{Neural Rendering}
Recent advances in computer graphics and vision \citep{liang2024photorealistic,liang2025diffusionrenderer,he2025unirelight} have seen diffusion models extend their utility beyond mere image generation. Notably, \cite{liang2024photorealistic} have explored using diffusion models as renderers for inverse rendering, demonstrating promising results, particularly in the domain of image-based object insertion. By framing the inverse rendering problem within the diffusion framework, these models can learn to synthesize realistic scene parameters that match input images.
Concurrent work by \cite{careaga2025physically} employs a generative pipeline, using a simplified diffuse renderer to guide a diffusion model. Our pipeline is the inverse: we use neural networks to initialize parameters for a physically based renderer. By enforcing the Disney BSDF and Monte Carlo ray tracing in the final output, we support complex material editing (roughness, metallic, transparency) with physically correct light transport, which are omitted or approximated by \cite{careaga2025physically}

\paragraph{Image and Material Editing}
Material editing generally follows two categories. The first involves inferring material properties through inverse rendering, then modifying them to achieve edits~\citep{li2018materials,azinovic2019inverse,zhu2022learning}. These methods provide strong interpretability, and using a physically based renderer makes the results highly reliable, though realism may sometimes be limited.
The second set is neural network-based. Following the success of DM~\citep{rombach2022high}, many image editing methods have emerged~\citep{mokady2023null,wang2024stereodiffusion,brooks2023instructpix2pix,shi2024dragdiffusion}. 
While recent diffusion-based editing methods~\citep{zeng2024rgb,lyu2025intrinsic} have shown impressive results in manipulating intrinsic components, they typically require edited 2D feature maps (e.g., normal and albedo maps) as input for object insertion. For general users, acquiring a physically consistent normal map for a new object is non-trivial and often requires rendering a 3D model first. In contrast, our work directly supports standard 3D mesh assets for insertion. This not only simplifies the workflow for users but also ensures physically accurate interactions, such as self-occlusion and casting shadows, which are challenging to achieve with 2D-layer-based manipulation.

Regarding material editing, Alchemist~\citep{sharma2024alchemist} is a representative example, trained on synthetic data to allow editing of albedo, roughness, metallic properties, and transparency. These methods benefit from the strengths of DM and achieve highly realistic results but have limited interpretability due to their neural network-based nature.
Our approach falls into the first category with a solid interpretability, meanwhile, it enables realistic results.
While early image-based methods \citep{khan2006image} simulated transparency via normal-based warping, we employ a geometry-aware approach that models light path traversal through the object volume ($d_1,d_2$), accounting for thickness-dependent refraction.

\paragraph{Light Estimation}
Light estimation is a distinct research field~\citep{park2020seeing,swedish2021objects,wang2022stylelight,tang2022estimating,yu2023accidental,dastjerdi2023everlight,verbin2024eclipse,enyo2024diffusion}. Some methods represent lighting implicitly~\citep{yao2022neilf,zhu2022learning,kocsis2024intrinsic}, limiting generalizability, while others require multi-view inputs and scene mesh data~\citep{park2020seeing,yu2023accidental,verbin2024eclipse}. Diffusion posterior illumination (DPI)~\citep{lyu2023diffusion} is related to our approach. This method combines differential rendering with DM for high-quality envmap generation but relies on multi-view NeRF methods for mesh reconstruction and lacks material BRDF data, reducing accuracy. In contrast, our approach uses MatNet's output for more accurate envmap optimization. % As our focus is not high-resolution envmaps, we limit resolution to $32\times16$, consistent with prior work \citep{li2021openrooms,choi2023mair}.

\begin{table}[!htb]
    \centering
    \caption{Summary of notations and symbols.}
    \label{tab:notation}
    \begin{tabular}
    {@{}p{0.2\linewidth}|p{0.75\linewidth}@{}}
    \toprule
    \textbf{Symbol} & \textbf{Description} \\ 
    \midrule
    \multicolumn{2}{c}{Geometry and Coordinates} \\
    \hline
    \rule{0pt}{2.5ex}$\bm{x} \in \mathbb{R}^3$ & Surface point position in world space \\
    $\bm{x}_s \in \mathbb{R}^2$ & Screen space coordinates corresponding to $\bm{x}$ \\
    $\mathbf{S}(\cdot)$ & Projection function mapping world position to screen space, $\bm{x}_s = \mathbf{S}(\bm{x})$ \\
    $\vec{N}(u,v)$ & Surface normal vector \\
    $\vec{n}$ & Surface normal vector at shading point \\
    $\vec{\omega}_i, \vec{\omega}_o$ & Incident light direction and outgoing view direction (normalized) \\
    
    $\vec{h}$ & Half vector, $\vec{h} = (\vec{\omega}_i + \vec{\omega}_o)/\|\vec{\omega}_i + \vec{\omega}_o\|$ \\
    $\Omega$ & Hemisphere centered at the surface normal \\
    $\mathbf{N}_\text{ndc}$ & Normalized Device Coordinates (NDC)\\
    
    \midrule
    \multicolumn{2}{c}{Material Properties} \\
    \hline
    \rule{0pt}{2.5ex}$\mathbf{A}, \mathbf{R}, \mathbf{M}$ & Albedo, Roughness, and Metallic maps \\
    $\mathbf{N}, \mathbf{D}$ & Normal map and Depth map \\
    $\mathbf{X}_p$ & Initial property predicted by MatNet (e.g., $\mathbf{A}_p, \mathbf{R}_p$) \\
    $\mathbf{X}^{\ast}$ & Final property after differentiable optimization (e.g., $\mathbf{A}^{\ast}$) \\
    $\eta$ & Index of Refraction (IOR) \\
    
    \midrule
    \multicolumn{2}{c}{Rendering and Lighting} \\
    \hline
    \rule{0pt}{2.5ex}$L_o(\bm{x}, \vec{\omega}_o)$ & Outgoing reflected radiance \\
    $L_i(\bm{x}, \vec{\omega}_i)$ & Incident radiance \\
    $L_{\mathbf{E}}$ & Radiance queried from the environment map \\
    $\mathbf{E}$ & Environment map (High Dynamic Range (HDR)
image) \\
    $f_r(\cdot)$ & Bidirectional Reflectance Distribution Function (BRDF) \\
    $V(\bm{x}, \vec{\omega}_i)$ & Visibility term (1 if visible, 0 if occluded) \\
    
    \midrule
    \multicolumn{2}{c}{Neural Networks and Optimization} \\
    \hline
    \rule{0pt}{2.5ex}$\mathrm{MatNet}(\cdot)$ & Material Prediction Network \\
    $\mathrm{MLP}_{\text{env}}$ & Multi-Layer Perceptron for environment map representation \\
    $\mathrm{MLP}_{\text{mat}}$ & Multi-Layer Perceptron for material property refinement \\
    $\Phi_{\text{env}}, \Phi_{\text{mat}}$ & Learnable weights (parameters) for the respective MLPs \\
    $\gamma(\cdot)$ & Positional encoding function applied to coordinates/features  \\
    $\mathbf{d}$& 3D unit direction vectors (input for $\mathrm{MLP}_{\text{env}}$) \\
    $\mathbf{m}$ & Concatenated material features used as input for $\mathrm{MLP}_{\text{mat}}$ \\
    $\mathcal{L}_{\text{re}}$ & Rendering reconstruction loss \\
    $\mathcal{L}_{\text{cons}}$ & Consistency constraint loss relative to MatNet predictions \\
    $\delta$ & Hyperparameter scaling the constraint loss $\mathcal{L}_{\text{cons}}$ \\
    $\zeta$ & Scaling factor for the output of the material MLP \\
    \bottomrule
    \end{tabular}
    
\end{table}

\paragraph{Shadows and Ray Tracing}
Shadows are crucial for photorealistic rendering. Previous work by \cite{li2022physically} emphasized shadow accuracy, using an OptiX-based ray tracer~\citep{parker2010optix} to compute shadows for neural network input. However, shadows are often neglected in recent research~\citep{zhu2022learning,yao2022neilf,li2023multi}; methods based on screen-space or image-based ray tracing lack geometric occlusion and thus fail to generate accurate shadows~\citep{yao2022neilf,kocsis2024intrinsic,zhu2022learning}. Many neural renderers struggle with shadow generation~\citep{choi2023mair,zhu2022irisformer}, as accurate shadows require physically based light transport, it is worth noting that this limitation has been utilized to identify AI-generated images~\citep{sarkar2024shadows}. % Our method includes full path tracing for physically accurate shadows. % (Fig. \ref{fig:render_comparison}).

Recent advances in DM-based neural rendering, notably the work presented in DiffusionRenderer \citep{liang2025diffusionrenderer}, demonstrate the capacity to generate reasonably realistic shadows without explicit 3D geometry. However, our approach, leveraging 3D geometry, consistently achieves more accurate shadow results compared to DiffusionRenderer. Furthermore, the incorporation of 3D geometry and ray tracing enables physically-based simulation within our framework, a capability currently challenging for diffusion model-based neural rendering techniques.

\section*{Nomenclature}
\label{sec:nomenclature}

To ensure clarity and consistency throughout the paper, we summarize the mathematical notations and symbols used in Table~\ref{tab:notation}.

\section{Preliminary} \label{sec:preliminary}
\subsection{Simplified DisneyBRDF}
In the following $\mathbf{A}$, $\mathbf{R}$, and $\mathbf{M}$ represent the image texture values of albedo, roughness, and metallic. To model surface scattering we use the DisneyBRDF~\citep{burley2012physically,burley2015extending} for our implementation. During the optimization process, we do not model sheen, clearcoat, and glass, so the BRDF in Eq.~\eqref{eq:render_eq} can be written
\begin{equation}
        f_r(\bm{x}, \vec{\omega}_i, \vec{\omega}_o)  = f_s +  (1-\mathbf{M}(\bm{x}_s)) f_d
        \label{eq:disneybrdf}
\end{equation}
with the specular ($f_s$) and diffuse ($f_d$) terms defined by
\begin{align}
    f_s &= \frac{F_s(\vec{\omega}_i, \vec{\omega}_o; \eta) D_s(\vec{h}; \mathbf{R}(\bm{x}_s)) G_s(\vec{\omega}_i, \vec{\omega_o}; \mathbf{R}(\bm{x}_s))}{4\,|\vec{n} \cdot \vec{\omega}_i| |\vec{n} \cdot \vec{\omega}_o|}  \\ 
    f_d &= \frac{\mathbf{A}(\bm{x}_s)}{\pi} F_d(\vec{\omega}_i) F_d(\vec{\omega}_o) \,,
\end{align}
where $F$ is the Fresnel reflectance, which here uses the half vector $\vec{h} = (\vec{\omega}_i + \vec{\omega}_o)/\|\vec{\omega}_i + \vec{\omega}_o\|$ and models the fraction of light reflected from the surface as a function of material properties, including the relative index of refraction~$\eta$. We use the modified Schlick approximation~\citep{schlick1994inexpensive,burley2012physically}:
\begin{align}
    \!F_s(\vec{\omega}_i, \vec{\omega}_o; \eta) &= C_0(\eta) + (1 - C_0(\eta))(1-\vec{h} \cdot \vec{\omega}_o)^5 \, \\ 
    C_0(\eta) &= R_0(\eta)(1-\mathbf{M}(\bm{x}_s))+\mathbf{M}(\bm{x}_s)\mathbf{A}(\bm{x}_s),
\end{align}
where $R_0(\eta)$ is the Fresnel reflectance of a dielectric at normal incidence. 
Similarly, for the diffuse term,
\begin{align}
F_d(\vec{\omega})& =1+(F_{D90}-1)(1-\vec{n}\cdot\vec{\omega})^5 \, \\
F_{D90} &=0.5+2\, \mathbf{R}(\bm{x}_s)\, (\vec{h}\cdot\vec{\omega})^2 \,.
\end{align}
Regarding the other terms, $D_s$ is the microfacet normal distribution function, while $G_s$ is the microfacet shadowing-masking function~\citep{schlick1994inexpensive}. We use the GGX distribution~\citep{Walter:ea:2007} for these terms due to its simplicity, with just one roughness parameter $\mathbf{R}(\bm{x}_s)$, while being able to fit empirical data well. 

\begin{figure*}[tb]
    \centering
    \includegraphics[width=0.9\linewidth]{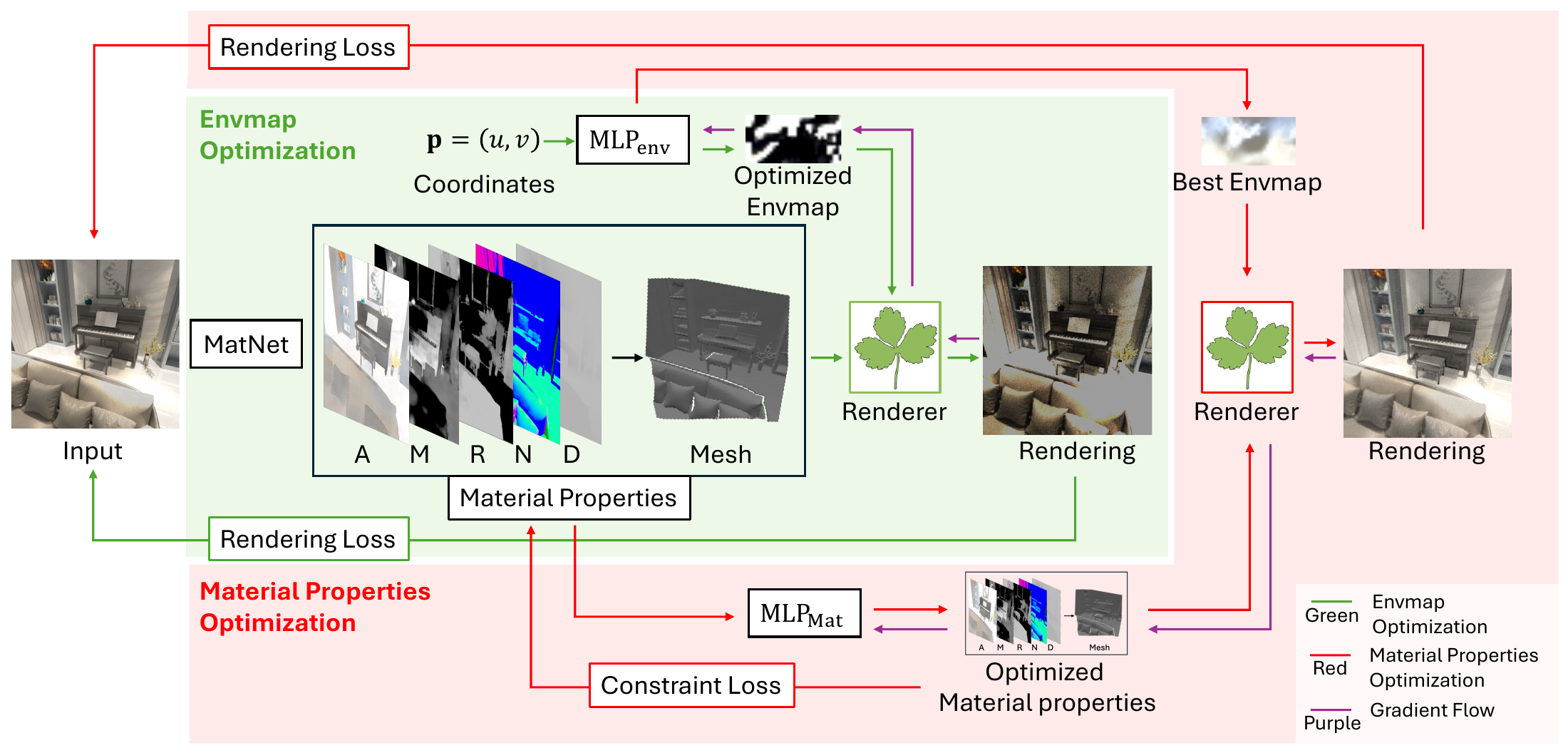}
    
    \caption{Our inverse rendering pipeline. Given an image, we use MatNet to predict material properties, followed by envmap optimization to estimate the lighting. We then perform material properties optimization, using the envmap that yields the smallest $\mathcal{L}_{\text{re}}$ during envmap optimization as the light source. The losses $\mathcal{L}_{\text{re}}$ and $\mathcal{L}_{\text{cons}}$ guide this process, allowing $\mathcal{L}_{\text{re}}$ to be minimized while keeping the results close to the MatNet predictions. See Sec.~\ref{sec:pdr} for further details.}
    
    \label{fig:pipeline}
\end{figure*}

\section{Methods} \label{sec:methods}
Our framework for single-view inverse rendering, material estimation, and editing follows a sequential pipeline, as illustrated in Fig.~\ref{fig:pipeline}. The process begins by inferring initial scene properties using a neural network. These properties are then progressively refined through a differentiable rendering optimization process to closely match the input image. The final optimized scene representation enables a variety of editing applications. Our pipeline consists of the following core stages:

\begin{enumerate}
    \item \textbf{Initial Property Prediction (Sec.~\ref{sec:matnet}):} We first employ a neural network, MatNet, to predict initial spatially-varying material properties (albedo, roughness, metallic), along with scene geometry (depth and normals) from a single input image.

    \item \textbf{Mesh Reconstruction (Sec.~\ref{sec:mesh_recon}):} Using the predicted depth map, we reconstruct a 3D triangle mesh of the scene. We introduce a novel boundary handling technique to ensure clean separation between foreground objects and the background, which is crucial for subsequent editing tasks.

    \item \textbf{Progressive Differentiable Rendering (Sec.~\ref{sec:pdr}):} As the initial MatNet predictions are not perfect, we introduce an iterative optimization framework. Using a physically based differentiable renderer, we progressively refine the environment map and material properties to minimize the reconstruction loss between the rendered image and the original input.
\end{enumerate}

This optimized scene representation serves as a foundation for high-quality applications such as relighting, object insertion, and physically based material editing, including a specialized technique for handling transparency (Sec.~\ref{sec:trans_edit}).

\subsection{Material Prediction Network} \label{sec:matnet}

To ensure semantic consistency and spatial alignment between geometry and material properties, we design a unified MatNet to jointly predict Albedo, Roughness, Metallic, Normal, and Depth. Unlike assembling multiple specialized off-the-shelf models (e.g., one for depth, another for decomposition), a unified architecture reduces domain gaps and simplifies the pipeline.

Due to the significant success of DM~\citep{rombach2022high} in image generation, recent work used it for the task of material property estimation, achieving impressive results~\citep{kocsis2024intrinsic}. However, this approach comes with high training costs and requires computationally intensive multi-step sampling during prediction. In comparison, we found that the model architecture for depth estimation \citep{depth_anything_v1} aligns more closely with the requirements of material prediction. We therefore adopt the dense prediction transformer (DPT) architecture~\citep{Ranftl2021}, which has been highly successful in depth estimation, for our material prediction task.

In contrast to \cite{zhu2022irisformer}, we utilize a pretrained DINOv2~\citep{oquab2023dinov2} encoder for feature extraction. Like previous research on depth estimation~\citep{birkl2023midas,depth_anything_v1}, we use two DPT decoders~\citep{Ranftl2021} for the regression of depth and material properties, respectively. Given the advancements in depth estimation, we initialize our training with weights from the pretrained depth model~\citep{depth_anything_v1} to expedite the training process. For the DPT depth decoder, we limit training to the last four layers of the RefineNet, while the weights of the DINOv2 encoder remain frozen throughout the training. The following loss function for $\mathcal{L}_{\text{MatNet}}$ is used to optimize the model,
\begin{equation}
    \mathcal{L}_{\text{MatNet}} = \sum_{i \in \{\mathbf{A}, \mathbf{R}, \mathbf{M}, \mathbf{N}, \mathbf{D}\}} \mathcal{L}_{i} \,,
\end{equation}
where the albedo loss $\mathcal{L}_{\mathbf{A}}$ combines the LPIPS perceptual loss and $L_1$ loss, while both the roughness loss $\mathcal{L}_{\mathbf{R}}$ and metallic loss $\mathcal{L}_{\mathbf{M}}$ are measured using the $L_1$ loss between the predictions and ground truth values. Finally, the $\mathcal{L}_{\mathbf{D}}$ is calculated using SiLogLoss~\citep{bhat2023zoedepth, eigen2014depth} to introduce logarithmic error and keep its scale constant to improve the robustness of depth prediction. This loss is defined by
\begin{equation}
    \mathcal{L}_{\mathbf{D}} = \sqrt{
        \begin{aligned}
            & \mathbb{E}_{i \in \text{mask}} \left( \left( \log(t_i) - \log(p_i) \right)^2 \right) \\
            & - \lambda \left( \mathbb{E}_{i \in \text{mask}} \left( \log(t_i) - \log(p_i) \right) \right)^2 \,,
        \end{aligned}
    } 
\end{equation}
where $t_i$ is the ground truth and $p_i$ is the predicted depth value for the valid pixel $i$, and we use $\lambda = 0.5$ as the hyperparameter that balances the variance and mean terms. To prevent outliers in the depth map from causing the model to collapse during training, we set the mask to a depth value in $[0,20]$. Finally, $\mathcal{L}_{\mathbf{N}}$ is calculated using cosine similarity.

The prediction process of the neural network can be expressed as
\begin{equation}
    \mathbf{A}_p,\mathbf{R}_p,\mathbf{M}_p,\mathbf{N}_p,\mathbf{D}_p = \text{MatNet}(I) \,,
    \label{eq:matnet}
\end{equation}
where $\mathbf{A}_p,\mathbf{R}_p,\mathbf{M}_p,\mathbf{N}_p,\mathbf{D}_p$ represent the albedo, roughness, metallic, normal (in linear space), and depth predicted by MatNet, and $I$ is the input image.

\subsection{Mesh Reconstruction} \label{sec:mesh_recon}

Using the predicted depth map $\mathbf{D}_p$, we project the 2D pixels into a 3D point cloud.  Standard meshing approaches \citep{Zhou2018,hu2021worldsheet} typically handle occlusion boundaries by simply discarding triangles that span large depth discontinuities. While effective for point cloud visualization, this approach leaves physical gaps (``white borders") between the foreground and background in a mesh. In inverse rendering, these gaps result in discontinuous light transport. 

\begin{figure}
    \centering
    \includegraphics[width=0.7\linewidth]{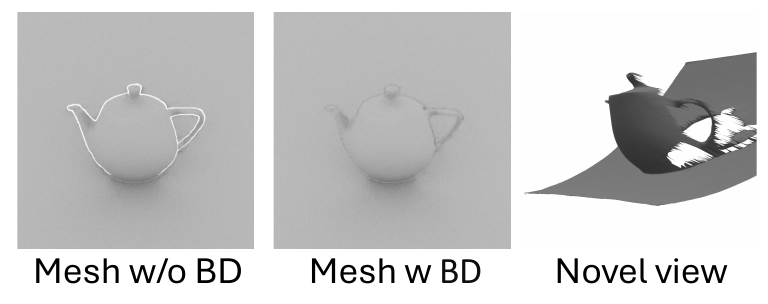}
    \caption{Our single-view mesh reconstruction method used on a test scene. Our method eliminates the white border at the edges. However, since our approach is based on the depth map, it primarily reconstructs the front view of the scene.}
    \label{fig:mesh_recon}
\end{figure}

To address this, we adopt a boundary handling strategy inspired by \cite{shade1998layered} and \cite{shih20203d}.
We identify boundary points by checking the grazing angle inequality: $\arcsin(\vec{n} \cdot \vec{v}) < \tau$, where $\vec{n}$ is the triangle's normal, $\vec{v}$ is the view direction, and $\tau$ is an angular threshold. Vertices satisfying this condition are marked as boundary points.
Instead of discarding the associated triangles, BD duplicates these boundary vertices: one set remains attached to the foreground geometry, while the duplicate set is projected to the depth of the nearest background neighbor. This effectively closes the gap with a back-facing surface, eliminating border artifacts (see Fig.~\ref{fig:mesh_recon}).

The threshold $\tau$ determines the sensitivity of the cut. Through ablation studies (see Supplementary Section \ref{sec:bd_ablation}), we determined that $\tau \approx 6^\circ$ provides an optimal trade-off between separating occlusion boundaries (avoiding under-segmentation) and preserving high-curvature surfaces (avoiding over-segmentation).
We also analyze failure cases in the supplementary material: specifically, structures thinner than the kernel size of the normal estimator ($\approx 7$ pixels) may fail to detach from the background, representing a resolution limit of our geometry reconstruction.

\subsection{Progressive Differentiable Rendering} \label{sec:pdr}

Our method assumes distant lighting, represented by an environment map (envmap), and does not explicitly model light sources within the scene. Ignoring emitters in the scene, the rendering equation becomes the reflected radiance equation:

\begin{equation}
    L_o(\bm{x}, \vec{\omega}_o) = \int_{\Omega} f_r(\bm{x}, \vec{\omega}_i, \vec{\omega}_o) L_i(\bm{x}, \vec{\omega}_i) (\vec{\omega}_i \cdot \vec{n}) \, d\omega_i \,,
\label{eq:render_eq}
\end{equation}
where  $L_o(\bm{x}, \vec{\omega}_o)$ is the outgoing radiance at the world space position $\bm{x}$ in the direction $\vec{\omega}_o$,
$f_r(\bm{x}, \vec{\omega}_i, \vec{\omega}_o)$ is the bidirectional reflectance distribution function (BRDF) at $\bm{x}$, representing the ratio of reflected radiance in the direction $\vec{\omega}_o$ to irradiance from the direction $\vec{\omega}_i$, while $L_i(\bm{x}, \vec{\omega}_i)$ is the incident radiance at $\bm{x}$ from the direction $\vec{\omega}_i$,
 $\vec{n}$ is the surface normal at $\bm{x}$, and
$\Omega$ is the hemisphere around $\vec{n}$ covering all directions of incidence. We use Simplified DisneyBRDF to model surface scattering. 
We let $\mathbf{S}(\bm{x}) = \bm{x}_s$ denote the transformation from world space to screen space coordinates. % Details of coordinate transformation and shading model are in Sec.~\ref{sec:preliminary}.
For the detailed mathematical formulation of the coordinate transformation between world space and screen space used in our rendering pipeline, please refer to the Supplementary Material.

\paragraph{Environment Map Optimization} \label{sec:env_opt}

Instead of directly optimizing the pixel values of the environment map $\mathbf{E}$, we represent the environment map implicitly using a Multi-Layer Perceptron (MLP). The input to the network determines the spatial consistency of the learned lighting.

To respect the spherical topology of the environment map and avoid discontinuities at the image boundaries ($u=0$ and $u=1$) or singularities at the poles, we map the normalized 2D equirectangular texture coordinates $(u, v) \in [0, 1]^2$ to 3D unit direction vectors $\mathbf{d} \in \mathbb{S}^2$:
\begin{equation}
    \mathbf{d} = (\sin\theta \cos\phi, \sin\theta \sin\phi, \cos\theta) \,,
\end{equation}
where $\phi = 2\pi u$ is the azimuth and $\theta = \pi v$ is the inclination.

We then apply sinusoidal positional encoding $\gamma(\cdot)$ to this 3D vector $\mathbf{d}$ to enable the network to capture high-frequency lighting details:
\begin{equation}
\begin{aligned}
\gamma(\mathbf{d}) = \bigl[
    &\sin(2^0 \pi \mathbf{d}), \cos(2^0 \pi \mathbf{d}), \dots, \\
    &\sin(2^{L-1} \pi \mathbf{d}), \cos(2^{L-1} \pi \mathbf{d})
\bigr] \, .
\end{aligned}
\end{equation}
Using 3D coordinates ensures that the learned radiance field is continuous and smooth over the entire spherical domain, providing better physical plausibility for distant lighting compared to raw 2D Cartesian coordinates.
where $L$ is the number of frequency bands. The MLP maps these encoded coordinates to RGB radiance values:
\begin{equation}
    \mathbf{E}(u, v) = \text{Softplus}(\mathrm{MLP}_{\text{env}}(\gamma(\mathbf{d}); \Phi_{\text{env}})) \,.
\end{equation}

We select \text{Softplus}$(x) = \ln(1+e^x)$ over bounded activations (e.g., sigmoid or tanh) because its unbounded range $(0,+\infty)$ strictly enforces non-negative radiance while preventing gradient saturation at high intensities. This linear asymptotic behavior is essential for accurately regressing high-value HDR components, such as sun disks.

In a differentiable rendering of the geometry $\mathbf{D}_p,\mathbf{N}_p$ based on Monte Carlo integration of Eq.~\eqref{eq:render_eq}, we evaluate the BRDF using $\mathbf{A}_p,\mathbf{R}_p,\mathbf{M}_p$ and the incident radiance using
\begin{equation}
L_i(\bm{x},\vec{\omega}_i) = V(\bm{x}, \vec{\omega}_i) L_{\mathbf{E}}(\vec{\omega}_i) \,,
\end{equation}
where $L_{\mathbf{E}}(\vec{\omega}_i)$ is the radiance from the envmap in the direction $\vec{\omega}_i$ and $V(\bm{x}, \vec{\omega}_i)$ is visibility, which is zero if geometry is found in the direction $\vec{\omega}_i$ and one otherwise.
To capture global lighting cues, the optimization is performed on the entire image domain without any masking.
Our optimization goal is to find the $\mathbf{E}^{\ast}$ that minimizes the difference between the rendering and the ground truth ($I$):
\begin{equation}
    \mathbf{E}^{\ast} = \text{argmin}_{\mathbf{E}}\mathcal{L_{\mathbf{E}}}(L_o(\bm{x},\vec{\omega}_o),I)
\end{equation}
where envmap loss $\mathcal{L_{\mathbf{E}}}$  combines $L_1$ and $L_2$ loss. The network weights $\Phi_{\text{env}}$ are discarded after optimization as we only need $\mathbf{E}^{\ast}$.

\paragraph{Material Properties Optimization} \label{sec:mat_opt}

Once we have an optimized environment map $\mathbf{E}^{\ast}$, we further refine the initial material properties predicted by MatNet. To strictly enforce the guidance from our neural prior, we formulate this optimization as learning a residual update to the initial predictions.

Let $\mathbf{m}_{p} = [\mathbf{A}_p, \mathbf{R}_p, \mathbf{M}_p]$ denote the initial material maps predicted by MatNet. We employ a position-embedded MLP to predict the residual offset
\begin{equation}
    \Delta \mathbf{m} = \mathrm{MLP}_{\text{mat}}(\gamma(\mathbf{m}_{p}); \Phi_{\text{mat}}) \,,
    \label{eq:mlp_residual}
\end{equation}
where $\gamma(\cdot)$ is the positional encoding, and $\Phi_{\text{mat}}$ denotes the learnable weights. The final optimized material properties $[\mathbf{A}, \mathbf{R}, \mathbf{M}]$ are computed by adding this residual to the initial prediction:
\begin{equation}
   [\mathbf{A}, \mathbf{R}, \mathbf{M}] = \mathbf{m}_{p} + \zeta \tanh(\Delta \mathbf{m}) \,,
   \label{eq:residual_sum}
\end{equation}
where $\tanh(\cdot)$ bounds the residual range to ensure stability, and $\zeta$ is a learnable scaling factor. This formulation ensures that the optimization starts from the physically plausible estimation provided by MatNet and refines it to match the rendering constraints.

The rendering process is the same as for the envmap, except that now we use $\bm{E}^{\ast}$ and keep it fixed.
Our optimization goal is to find $\mathbf{A}^{\ast}, \mathbf{R}^{\ast}, \mathbf{M}^{\ast}$ that minimize the difference between the rendered result and the ground truth ($I$):
\begin{equation}
    \mathbf{A}^{\ast}, \mathbf{R}^{\ast}, \mathbf{M}^{\ast} = \text{argmin}_{\mathbf{A,R,M}} \mathcal{L}_{\text{Mat}}(L_o(\bm{x},\vec{\omega}_o), I)
\end{equation}
where $\mathcal{L}_{\text{Mat}}$ is the material loss:
\begin{align}
\mathcal{L}_{\text{Mat}} &= \mathcal{L}_{\text{re}} + \delta \mathcal{L}_{\text{cons}} \,, 
\label{eq:cons_loss}
\end{align}
and the individual terms are:
\begin{align}
\mathcal{L}_{\text{re}} &= \mathcal{L}_{L1}(L_o, I) + \mathcal{L}_{L2}(L_o, I) \\
\mathcal{L}_{\text{cons}} &= \sum_{X \in \{\mathbf{A}, \mathbf{R}, \mathbf{M}\}} \mathcal{L}_{L1}(X_o, X_p) \,,
\end{align}
where $\mathcal{L}_{L1}$ and $\mathcal{L}_{L2}$ represent the $L_1$ and $L_2$ losses, respectively, and $\mathcal{L}_{\text{re}}$ measures the rendering loss between $L_o$ and $I$. The constraint loss $\mathcal{L}_{\text{cons}}$ compares the optimized and predicted material properties, introducing constraints to align the optimization with MatNet's predictions while minimizing the difference between $L_o$ and $I$. The scaling factor $ \delta $ controls the deviation from MatNet predictions: a smaller $ \delta $ allows more flexibility, while a larger $\delta$ keeps the optimization closer to MatNet, potentially increasing the gap between the rendering and the target.

\subsection{Opaque Material Editing} \label{sec:opak_mat_edit}
For $\mathbf{R, M, A}$ edits (excluding transparency), we use the SAM2 \citep{ravi2024sam2} model to segment the input image, creating a mask for the edit region. 
Note that SAM2 is employed exclusively for the editing task to isolate the target object selected by the user. It is not involved in the previous environment map estimation or material optimization steps.
We can then modify values within this region and re-render an image. For $\mathbf{A}$ edits, we convert to HSV color space for intuitive color adjustments.

\paragraph{Material Transparency Editing} \label{sec:trans_edit}
As described in Sec.~\ref{sec:mesh_recon}, we reconstruct the mesh using the depth map, capturing only the visible front side of objects (see Fig.~\ref{fig:mesh_recon}). This limits transparent object editing, as refractive effects cannot be fully modeled. While existing methods generate multi-view images for mesh reconstruction via NeRF~\citep{mildenhall2021nerf,liu2023syncdreamer,long2024wonder3d,shi2023zero123plus} or Gaussian Splatting \citep{kerbl20233d,guedon2024sugar,yu2024mip}, 
we propose a method to approximate material transparency editing within a single-view constraint, avoiding full scene reconstruction.
We model transparent objects using the following variant of the DisneyBSDF~\citep{burley2015extending}:
\begin{equation}
    f_\text{glass} = \frac{\sqrt{\mathbf{A}_\text{glass}(\bm{x}_s)}(1-F_s) D_s G_s |\vec{h} \cdot \vec{\omega}_o| |\vec{h} \cdot \vec{\omega}_i|}{|\vec{n} \cdot \vec{\omega}_i | |\vec{n} \cdot \vec{\omega}_o| (\vec{h} \cdot \vec{\omega}_i+\eta\, \vec{h} \cdot \vec{\omega}_o )^2} \,,
\end{equation}
where $\mathbf{A}_\text{glass}(\bm{x}_s) = (1-\mathbf{M}(\bm{x}_s)) \mathbf{A}_\text{BG}(\bm{x}_s) T$, with $T$ denoting specular transmission and $\mathbf{A}_\text{BG}$ the albedo of the diffuse background object after refraction (distinct from $\mathbf{A}$). Thus, by knowing the refraction points, we can simulate transparency effects without a back-side mesh, supporting single-view transparency editing.
It is important to note that our system first inpaints the albedo in the occluded region using a standard inpainting network and then uses this inpainted albedo as a texture for the underlying geometry during rendering. 

We use SAM2 \citep{ravi2024sam2} for object mask extraction and an inpainting model \citep{rombach2022high} to generate $ \mathbf{A}_\text{BG}$. For an incoming ray of direction $-\vec{\omega}_o$ hitting the masked area, the direction of the transmitted ray $\vec{\omega}_t$ is computed using the law of refraction~\citep{kay1979transparency}. 
Assuming the object's back normal is the same as its front normal $\mathbf{N}_p(\bm{x}_s)$, the world position after two refractions is
\begin{equation}
\bm{x}_2 = \bm{x} + d_1 \, \vec{\omega}_{t1} + d_2 \, \vec{\omega}_{t2} \,,
\end{equation}
where $\bm{x}$ is the initial position of incidence, $d_1$ and $d_2$ are refraction distances, and $\vec{\omega}_{t1}$ and $\vec{\omega}_{t2}$ are the directions after first and second refractions, respectively. Note that $\vec{\omega}_{t2}$ depends on $\mathbf{N}_p(\mathbf{S}(\bm{x} + d_1\,\vec{\omega}_{t1}))$. We trained a neural network on synthetic data to predict $d_1$ and $d_2$ for various shapes. %; see the supplementary materials for network details.
Thus, $\mathbf{A}_\text{BG}(\bm{x}_s) = \mathbf{A}_\text{BG}(\mathbf{S}(\bm{x}_2))$, 
where $\mathbf{S}(\bm{x}_2)$ are the screen coordinates of $\bm{x}_2$ after double refraction.
Regarding the refraction length, we made a simplifying assumption. We assumed the edited object was symmetrical along its front-to-back axis, and therefore calculated the refraction distance based on the depth predicted by MatNet.

\textbf{Thickness Prior Prediction.}
Simulating physically accurate refraction requires the exit point of the light ray, which depends on the object's thickness and back-face geometry. Since the back-face is occluded in a single view, we formulate this as a learning problem. We train a U-Net based RefractionNet same as the depth map prediction module, to predict a thickness map from the input image. 
Our network learns semantic shape priors (e.g., recognizing that a cup's handle is thinner than its body), providing a more plausible approximation for the ray travel distance. 
Our sensitivity analysis (see Suppl. Sec. \ref{sec:sensitivity_d}) demonstrates that editing results are perceptually robust to thickness prediction errors up to 50\%, validating the feasibility of this approximation.

\textbf{Background Inpainting}
Refraction reveals the background occluded by the opaque object. We utilize a inpainting pipeline \citep{rombach2022high,wu2025qwenimagetechnicalreport} to inpaint the background. The final color is queried using the screen-space coordinates of the ray after exiting the object.

\begin{table*}[!htb]
    \caption{Quantitative evaluation of material prediction on the InteriorVerse synthetic dataset~\citep{zhu2022learning}. We compare our MatNet against recent state-of-the-art methods across standard metrics for albedo, roughness, metallic, depth, and normal estimation. Best results are in \textbf{bold}, and second best results are \underline{underlined}.  DiffusionRenderer \citep{liang2025diffusionrenderer} trained on a more diverse dataset, which is likely the reason for the slightly larger deviation from GT on this dataset. Concurrent work by \cite{ke2025marigold} demonstrates strong performance for albedo and normals on this dataset.
    }
    \centering
    \resizebox{\linewidth}{!}{
    \begin{tabular}{c|ccc|ccc|ccc|cc|c}
    \toprule
         &  & Albedo & \textbf{} & & Roughness & \textbf{} &  & Metallic & \textbf{} & \multicolumn{2}{c|}{Depth} & Normal \\ 
        ~ & PSNR$\uparrow$ & SSIM$\uparrow$ & LPIPS$\downarrow$ & PSNR$\uparrow$ & SSIM$\uparrow$ & LPIPS$\downarrow$ & PSNR$\uparrow$ & SSIM$\uparrow$ & LPIPS$\downarrow $& MAE$\downarrow$ & RMSE$\downarrow$ & Angular Error$\downarrow$  \\ \hline
        \rule{0pt}{2.5ex}\cite{li2022physically} & 6.173 & 0.488 & 0.529 & 8.665 & 0.529 & 0.559 & - & - & - & - & - & 37.696$^\circ$  \\ 
        \cite{zhu2022learning} & 13.407 & 0.680 & 0.302 & 13.670 &\underline{0.596} & 0.360 & 15.600 & 0.489 & 0.281 & 1.828 & 2.045 & 26.872$^\circ$  \\ 
        \cite{zeng2024rgb} &18.149&0.791&0.183&\underline{15.926}&0.591&\underline{0.296}&14.078&0.642&0.273&-&-&14.155$^\circ$\\
        \cite{kocsis2024intrinsic} & 12.774 & 0.656 & 0.307 & 9.070 & 0.516 & 0.384 & 7.228&0.205&0.375 &- & - &-\\
         \cite{ke2025marigold} & \underline{19.223} & \textbf{0.821} & \underline{0.143} & 14.254 & 0.504 & 0.299 & 14.073 &0.657&0.296 & \underline{0.532} & \underline{0.958}&\textbf{10.052}$^\circ$\\
        \cite{liang2025diffusionrenderer} & 18.446 & 0.802 & 0.190 & 14.956 & 0.529 & \textbf{0.294} & \underline{14.477}&\underline{0.660}&\underline{0.280} &- & - &23.734$^\circ$\\
        Ours & \textbf{19.530} & \underline{0.811} & \textbf{0.108} & \textbf{16.632} & \textbf{0.663} & \textbf{0.294} & \textbf{18.660} & \textbf{0.701} & \textbf{0.243} & \textbf{0.244} &\textbf{ 0.491} & \underline{11.359$^\circ$} \\ 
    \bottomrule 
    
    \end{tabular}}
    \label{tab:mat_pred}
    \vspace{-1ex}
\end{table*}

\begin{figure*}[!htb]
    \centering
    \includegraphics[width=\linewidth]{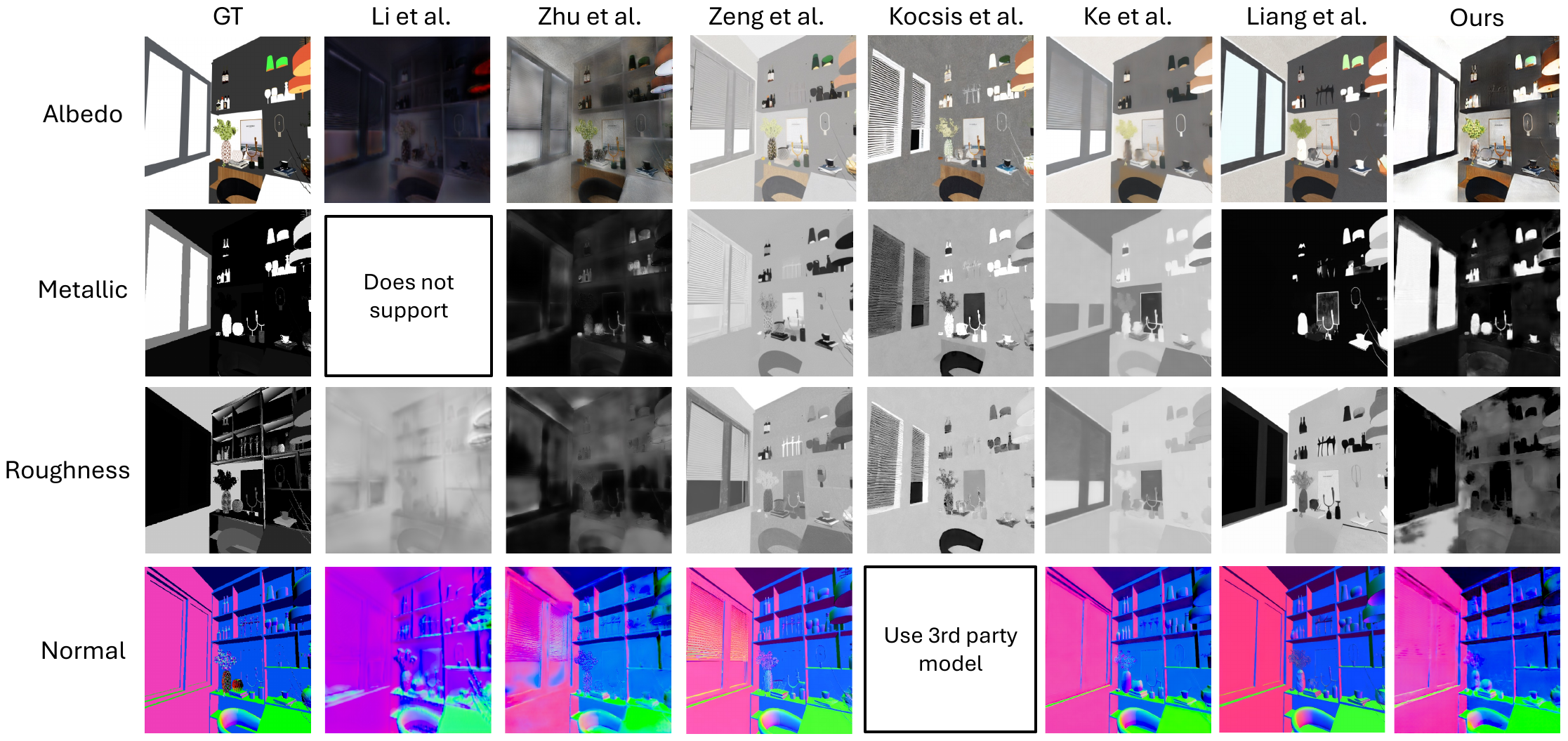}
    \caption{Qualitative results of material prediction on synthetic dataset. All normal maps are visualized in $[0, 1]$ RGB space, and linear albedo predictions are gamma corrected to sRGB.}
    \label{fig:bm_mat_pred_syn}
\end{figure*}

\begin{figure*}[!tb]
    \centering
    \includegraphics[width=0.9\linewidth]{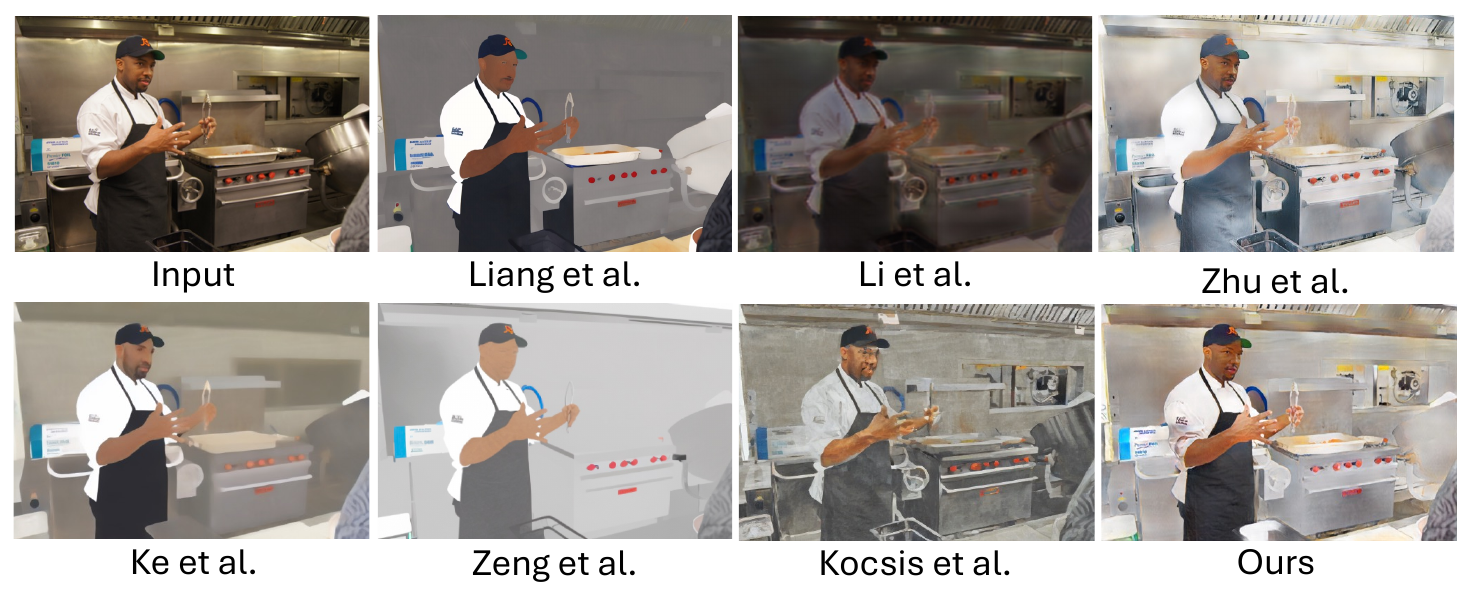}
    \caption{Qualitative results of albedo prediction on the IIW dataset~\citep{bell14intrinsic} highlight key differences: \cite{zhu2022learning} retain excessive details, \cite{kocsis2024intrinsic} distort textures and struggle with facial features, while  \cite{zeng2024rgb} over-smooth, losing critical details like facial features. These issues underscore the challenge of balancing detail preservation and realistic material prediction. Concurrent work by \cite{liang2025diffusionrenderer} and \cite{ke2025marigold} demonstrates a more balanced performance.}
    \label{fig:bm_mat_pred_iiw}
\end{figure*}

\section{Experiments} \label{sec:experiments}

It is important to note that MatNet serves as a support component within our pipeline; we do not directly compete with image decomposition works. Rather, more precise material predictions enhance the optimization of envmaps through the utilization of our pipeline.

\subsection{Implementation Details} \label{sec:implement_details}
\paragraph{MatNet}

MatNet is trained on the InteriorVerse dataset \citep{zhu2022learning}, following the official train/validation/test splits provided by the dataset to ensure consistency. The training set contains 45,073 images, while the validation set includes 5,024 images. Results are reported on a test set of 2,672 samples. We apply only random vertical flipping and random 512$\times$512 cropping as data augmentation. The model employs the DINOv2-ViT-B/14 encoder and a DPT decoder. It comprises 106.7 million parameters and requires 202.0 GFLOPs for inference. The model is trained on one A100 (80GB) for 7 days, optimized using AdamW with a learning rate of $10^{-5}$.

\paragraph{Progressive Differentiable Rendering}
We implement our pipeline using Mitsuba 3 \citep{jakob2022mitsuba3}. Gradients are computed using the Path Replay Backpropagation (PRB) integrator, which provides a memory-efficient solution for differentiating path-traced images. We use 64 samples per pixel (SPP) for optimization. Notably, we do not apply any denoiser during the optimization process to prevent the introduction of bias or artifacts into the gradients.
We assume a pinhole camera model with a 35 degrees field of view (FOV), 512$\times$512 image resolution, and $3$ color channels. 
Optimization of progressive differentiable rendering can be in stages: first optimizing $\mathbf{R}$ and $\mathbf{M}$, then $\mathbf{A}$. As albedo has the most significant impact on the rendering, increasing its scale factor in $\mathcal{L}_{\text{cons}}$ during joint optimization prevents the network from overfocusing on albedo, ensuring a balanced material property optimization.
All optimization results are reported from single runs with a fixed random seed to ensure deterministic reproducibility.

\paragraph{Default Automatic Optimization Schedule}
To ensure robust reconstruction for ``in-the-wild'' images without manual tuning, we propose a two-stage optimization strategy:
\begin{enumerate}
    \item \textbf{Initialization:} 
    Directly using raw predictions (which may contain values near 0 or 1) can lead to unstable optimization (e.g., vanishing or exploding gradients). We therefore remap the initial roughness ($\mathbf{R}_p$) and metallic ($\mathbf{M}_p$) predictions from MatNet to a more optimizable range using the transformation $x' = 0.7x + 0.2$. Albedo ($\mathbf{A}_p$) is initialized directly from MatNet predictions as it is generally more reliable. $\delta$ in Eq.~\eqref{eq:cons_loss} is set to 0.5.
    
    \item \textbf{Stage 1 (Roughness \& Metallic):} 
    Since albedo has the most dominant effect on the rendering loss, simultaneously optimizing all parameters often leads to the ``albedo shortcut" problem, where lighting/material errors are baked into the albedo. To mitigate this, we freeze the albedo and strictly optimize $\mathbf{R}$ and $\mathbf{M}$ in the first stage.
    
    \item \textbf{Stage 2 (Joint Refinement):} 
    Once $\mathbf{R}$ and $\mathbf{M}$ have stabilized, we unlock $\mathbf{A}$ and perform a joint optimization of all material properties to fine-tune the reconstruction.
\end{enumerate}
A supplementary video demonstrates this multi-stage optimization process.

\paragraph{Hyperparameters and Timing} \label{sec:hypermeters}
We use the Adam optimizer for both stages. 
For material properties, the learning rate is set to $3 \times 10^{-4}$ with a StepLR scheduler (step size 100, $\gamma=0.8$). 
For the environment map, the learning rate is $10^{-3}$ with the same scheduler.
To ensure efficiency, we implement an early stopping mechanism: the optimization terminates if the loss improvement is less than 1\% for 100 steps for the envmap, and 0.5\% for 200 steps for materials.
The entire optimization process for a single real-world image typically converges within 3 to 7 minutes on a single NVIDIA RTX 3090 GPU.

\subsection{Material Prediction Accuracy} \label{sec:exp_mat_pred}

\paragraph{Synthetic Dataset}
We evaluate the performance of our MatNet on the test set of the InteriorVerse \citep{zhu2022learning} synthetic indoor dataset. %As shown in Table \ref{tab:mat_pred}, while the very recent concurrent work, DiffusionRenderer \cite{liang2025diffusionrenderer}, has achieved state-of-the-art results in overall performance, our method consistently secures the second-best performance across most evaluated metrics. Crucially, our MatNet stands out with the fastest inference speed.
As seen in Table~\ref{tab:mat_pred}, our method achieves the best performance on nearly all metrics. Qualitative comparisons are shown in Fig.~\ref{fig:bm_mat_pred_syn}. It is worth noting that the concurrent work, DiffusionRenderer \citep{liang2025diffusionrenderer}, also performs well. While its quantitative scores are slightly lower on some metrics, which may be due to its training on a more diverse dataset leading to stylistic differences from the ground truth, its qualitative results in Fig.~\ref{fig:bm_mat_pred_syn} reveal perceptually clean metallic and roughness maps.

It is important to note that our approach is not in conflict with DiffusionRenderer, it offers a complementary advantage. We posit that our method's more accurate material property predictions can provide excellent initial conditions for optimization, which in turn significantly accelerates progressive differentiable rendering, potentially enhancing the overall efficiency and quality of such advanced rendering pipelines.

\begin{table}[!tb]
    \caption{Quantitative evaluation of albedo predictions on the IIW Dataset~\citep{bell14intrinsic} using WHDR (with $\delta=0.1$).  We include comparisons with concurrent work by \cite{ke2025marigold} and \cite{liang2025diffusionrenderer}. While \cite{ke2025marigold} achieve the lowest WHDR, our method delivers the faster inference speed, offering a compelling trade-off between efficiency and performance. \textit{Time} denotes the average inference time in seconds per image on one H100 (80G). It is important to emphasize that the reported time reflects only the inference phase of MatNet and does not account for any subsequent optimization procedures.}
    \label{tab:whdr}
    \centering
    \captionsetup{width=\linewidth} 
    \begin{tabular}{c|c|cc}
    \toprule
         \multirow{2}{*}{Methods}& \multirow{2}{*}{WHDR $\downarrow$} & \multicolumn{2}{c}{Time ($s$) $\downarrow$} \\
         &  & Avg & Med \\
         \hline
        \rule{0pt}{2.5ex}\cite{li2022physically} & 0.342 & \underline{0.020}  & \underline{0.020} \\ 
        \cite{zhu2022learning} & 0.232 & 0.025  & 0.025 \\ 
        \cite{zeng2024rgb} & 0.190 & 7.753  & 7.631 \\
        \cite{kocsis2024intrinsic} & 0.206 & 6.817  & 6.395 \\ 
        \cite{liang2025diffusionrenderer} & \underline{0.178} & 6.205 & 6.023 \\
        \cite{ke2025marigold} & \textbf{0.147} & 0.530 & 0.530 \\
        Ours & 0.197 & \textbf{0.018} & \textbf{0.018} \\ 
        \bottomrule
    \end{tabular}
\end{table}

\begin{table*}[!tb]
    \centering
    \setlength{\tabcolsep}{1mm}
    \caption{Quantitative evaluation of envmap optimization on the constructed dataset. 
    We compare with \cite{phongthawee2024diffusionlight} and \cite{lyu2023diffusion}. 
    Our method is evaluated at a resolution of $512\times256$ using spherical coordinates.
    \textit{Envmap} metrics measure pixelwise accuracy, while \textit{Render} metrics assess the relighting quality.
    SH Error measures the discrepancy in Spherical Harmonics (SH) coefficients, serving as a robust proxy for lighting direction and color accuracy. ``w/ mat'' and ``w/o mat'' refer to whether fixed material properties predicted by MatNet are provided during the envmap optimization.
    Our method (w/ mat) achieves lower SH Error, indicating better physical plausibility.
    }   
    \begin{tabular}{c|cc|cc|cc|c|c}
    \toprule
        ~ & \multicolumn{2}{c|}{SSIM $\uparrow$} & \multicolumn{2}{c|}{PSNR$\uparrow$} & \multicolumn{2}{c|}{MSE $\downarrow$} & SH Error $\downarrow$& Time $\downarrow$ \\ 
        ~ & Envmap & Render & Envmap & Render & Envmap & Render & Envmap & ($s$) \\ \hline
        \cite{phongthawee2024diffusionlight} & 0.316 & 0.454 & 8.874 & 14.419 & 1.302 & 0.108 & 3.382 & \textbf{61}  \\ 
        \cite{lyu2023diffusion} & 0.286 & 0.598 & 9.691 & 17.359 & 1.244 & 0.093 & 3.178 & 453  \\  \hline
        Ours w/o mat & 0.206 & 0.645 & 9.890 & 17.300 & 1.688 & 0.192 & 4.554 & 112  \\ 
        \textbf{Ours w/ mat} & \textbf{0.520} & \textbf{0.811} & \textbf{17.175} & \textbf{20.048} & \textbf{0.791} & \textbf{0.029} & \textbf{0.474} & 105  \\
    \bottomrule
    \end{tabular}
    \label{tab:envmap_bm}
\end{table*}

\paragraph{Real World Dataset}

We evaluate albedo predictions on the IIW dataset using the Weighted Human Disagreement Rate (WHDR), as proposed by \cite{bell14intrinsic}, which measures errors against human annotations. Qualitative results are provided in Fig.~\ref{fig:bm_mat_pred_iiw}.

We also conducted rigorous inference speed testing for different models. To avoid disk I/O interference caused by image reading/writing, we used randomly generated noise as the input to the model. The experiments were performed on a single H100 (80GB) GPU with a batch size of 1. The model was first warmed up by running 50 iterations (the results from this phase were excluded from the final statistics). After the warm-up, the model was executed for 500 iterations, and both the average and median inference times were calculated. Results are shown in Table~\ref{tab:whdr}.

\begin{figure*}
    \centering
    \includegraphics[width=\linewidth]{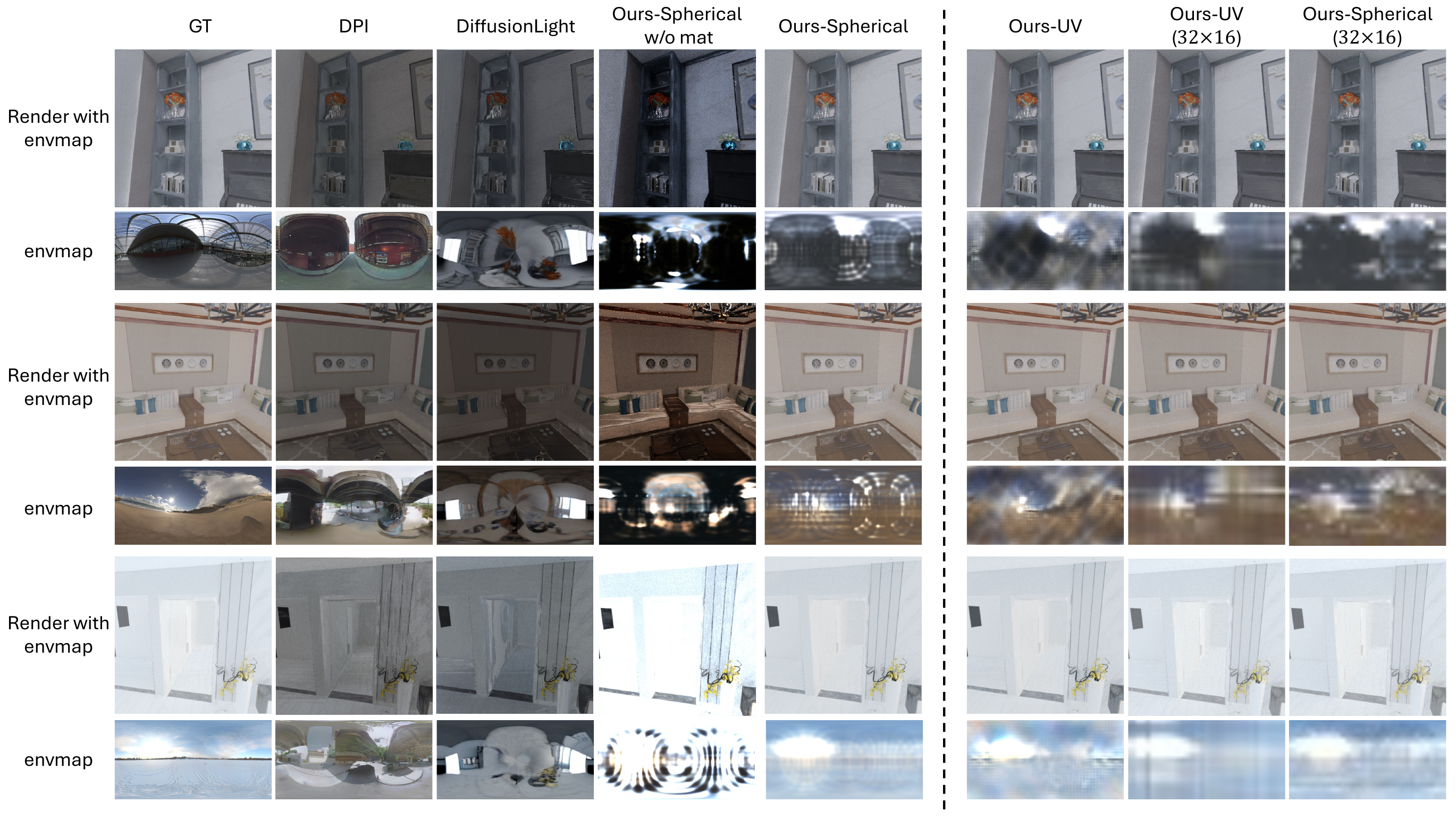}
    \caption{Qualitative results of envmap optimization. The left part of the figure presents a qualitative comparison of results across various methods, with our approach demonstrating competitive performance. For instance, our method accurately reproduces the reflections on the vase in the first row, while the optimized envmap correctly reconstructs the luminous central region. The right part presents the ablation study results with different coordinate systems and envmap resolutions. Although the differences in the rendered images are subtle, the optimized envmap exhibits significant variations (cf.~Sec.~\ref{sec:ablation_env}).}
    \label{fig:env_bm}
\end{figure*}

\begin{figure*}
    \centering
    \includegraphics[width=0.8\linewidth]{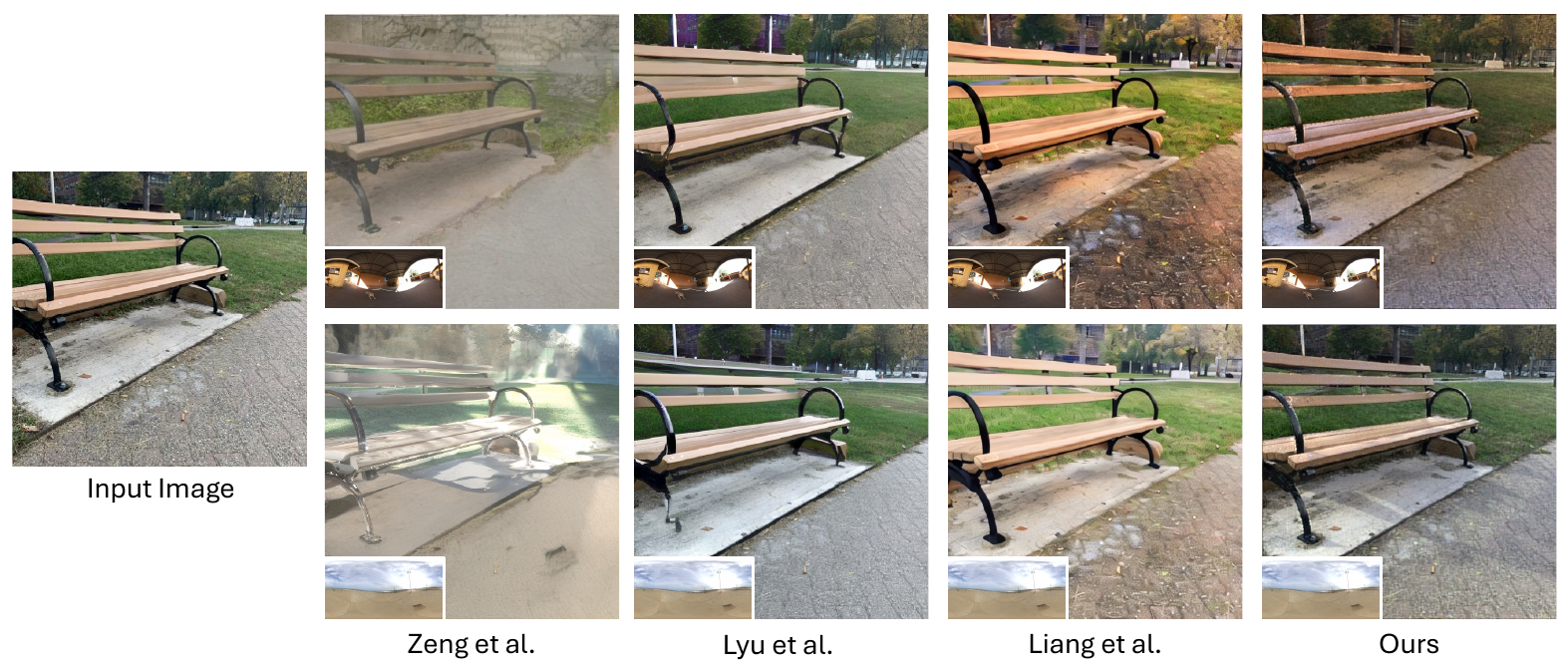}
    \caption{Relighting comparison. The lower left corner of the images shows the envmap used for relighting. The results produced by IntrinsicEdit~\citep{lyu2025intrinsic} and RGB$\leftrightarrow$X \citep{zeng2024rgb} may have certain discrepancies in performance, as they originally accept irradiance maps rather than envmaps as input for relighting. We approximate the required input irradiance map by sampling the envmap according to surface normal directions. DiffusionRenderer \citep{liang2025diffusionrenderer} was originally designed to support envmaps and produce more resonable results, however it still fails to correctly ``render" shadows (2nd row).}
    \label{fig:bench_relit}
\end{figure*}

\begin{figure*}
    \centering
    \includegraphics[width=0.8\linewidth]{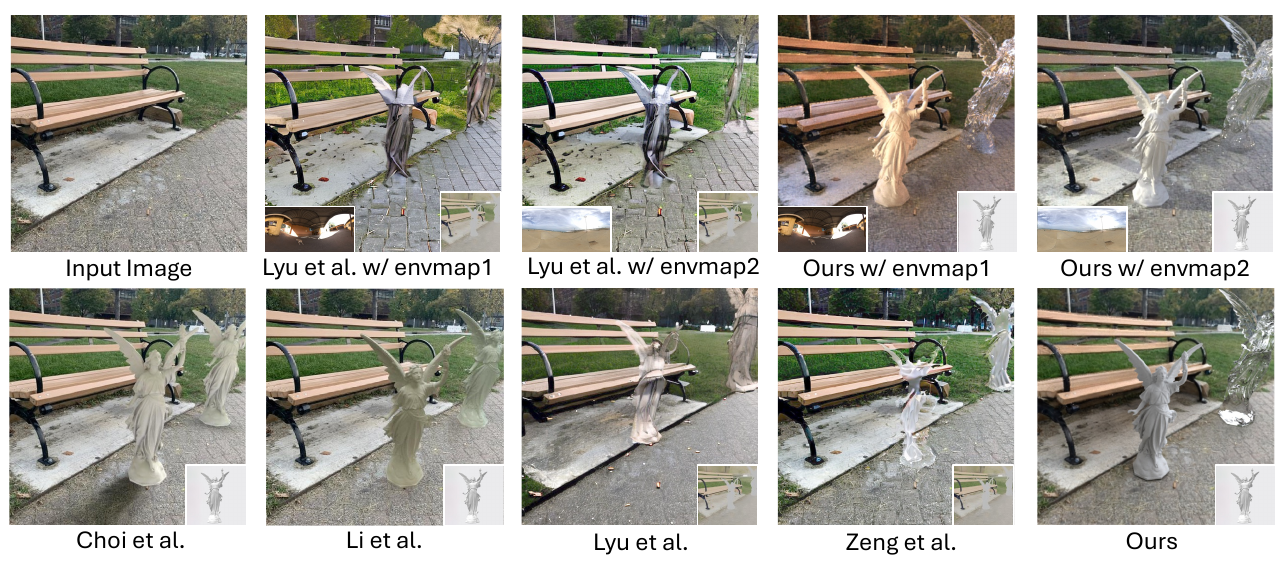}
    \caption{
    Object insertion qualitative comparison. The first row presents the results of object inversion combined with relighting using a new envmap, while the second row shows the results of object inversion without changing the lighting conditions. The bottom-right corner of the image indicates the type of object inversion. \cite{zeng2024rgb} and \cite{lyu2025intrinsic} employ an edited g-buffer (we manually composed rendered albedo and normal maps into the image, only the albedo is shown in the figure), whereas other methods utilize a mesh.
    Our method naturally handles geometric occlusion and generates more realistic shadows (see the contact region between the object and the floor) via physics-based rendering, whereas the 2D-based method lacks explicit geometry to compute accurate light transport.}
    \label{fig:oi_qual}
\end{figure*}

\subsection{Environment Map Optimization} \label{sec:exp_env_opt}

Currently, single-view envmap generation primarily relies on DM \citep{phongthawee2024diffusionlight}, which can only produce visually plausible envmaps. Optimization-based methods often require multi-view input~\citep{park2020seeing,yu2023accidental,verbin2024eclipse} or use implicit lighting representations~\citep{yao2022neilf,zhu2022learning,kocsis2024intrinsic}. For optimization approaches, we selected DPI~\citep{lyu2023diffusion} as our benchmark, adapting it to single-view input and providing it with a mesh reconstructed using our method (Sec.~\ref{sec:mesh_recon}).

We evaluate the performance using a constructed dataset containing 1,000 evaluation samples, derived from 10 diverse scenes, each rendered from 10 viewpoints under 10 distinct lighting conditions. Our envmap is optimized at a resolution of $512 \times 256$ (width$\times$height) using spherical coordinates. In the quantitative evaluation, the notation ``w/ mat'' refers to using the material properties (Albedo, Roughness, Metallic) predicted by MatNet as fixed priors during the environment map optimization.. Experiments were carried out on a single RTX 3090. Our method employs a StepLR scheduler with a step size of 100 and a decay factor of ($\gamma = 0.8$), and the learning rate is set to ($10^{-3}$). Early stopping is also applied: the optimization terminates if the loss improvement is less than 1\% for 100 consecutive steps, with a maximum of 500 optimization steps.

Evaluation metrics included the difference between the optimized and the ground truth envmaps and the difference between relit scenes using the optimized and the ground truth envmaps. Both methods used ground truth material properties for relighting. 

As shown in Table~\ref{tab:envmap_bm}, our method outperforms baselines across key metrics.
Most notably, we achieve a drastically lower SH Error ($0.474$) compared to DiffusionLight ($3.382$) and DPI ($3.178$). 
Since Spherical Harmonics (SH) represent the low-frequency global structure of lighting, this indicates that our physically based differentiable rendering pipeline recovers the lighting direction and color temperature more accurately than learning-based or implicit methods.
Although DiffusionLight~\citep{phongthawee2024diffusionlight} is faster (61\,s), it struggles with physical consistency. Our method strikes a balance, being faster than the comparable optimization method DPI (105\,s vs. 453\,s) while delivering better quality (Rerender PSNR $20.05$ vs. $17.36$).
Qualitative results are in Fig.~\ref{fig:env_bm}.
Notably, while rerendering error is low, envmap error remains relatively high due to the inherent ambiguity in single-view optimization; accurate lighting in certain regions can yield a close match to the ground truth, even if the overall envmap differs significantly.

\subsection{Object insertion and relighting} \label{sec:oi}

We compare our method with recent editing methods \citep{lyu2025intrinsic,liang2025diffusionrenderer,zeng2024rgb}. Note that IntrinsicEdit~\cite{lyu2025intrinsic} and RGX$\leftrightarrow$X \citep{zeng2024rgb} operate on 2D intrinsic layers and utilize irradiance maps rather than environment maps. We adapted the inputs as follows to facilitate a fair comparison:

\begin{enumerate}
\item \textbf{Irradiance Approximation:} \cite{zeng2024rgb} and \cite{lyu2025intrinsic} do not natively support high frequency environment maps. We approximated the required input irradiance map by sampling the envmap $\mathbf{E}$ based on the surface normal directions $\vec{N}$. Specifically, for each pixel $(u,v)$, we map the normal vector $\vec{N}(u,v)$ to spherical coordinates $(\theta,\phi)$ and sample the corresponding value from $\mathbf{E}$. Note that this is a simplified approximation that ignores self-shadowing and global integration.
\item \textbf{2D Layer Generation:} For the object insertion task, \cite{zeng2024rgb} and \cite{lyu2025intrinsic} require an edited albedo and normal map. We generated these by rendering our 3D mesh assets under constant ambient lighting to produce the corresponding 2D layers, which were then pasted into the scene's intrinsic maps.
\end{enumerate}

As shown in Figs.~\ref{fig:bench_relit} and \ref{fig:oi_qual}, while \cite{lyu2025intrinsic} generate visually harmonious results, our method produces significantly more realistic lighting effects. Specifically, our physically based ray tracing accurately captures cast shadows from the inserted object onto the scene (and vice versa) and preserves high-frequency specular reflections and refraction of the transparent object, which are impossible in the irradiance-based approximation of 2D methods.

\begin{figure}[!tb]
    \centering
    \includegraphics[width=0.9\linewidth]{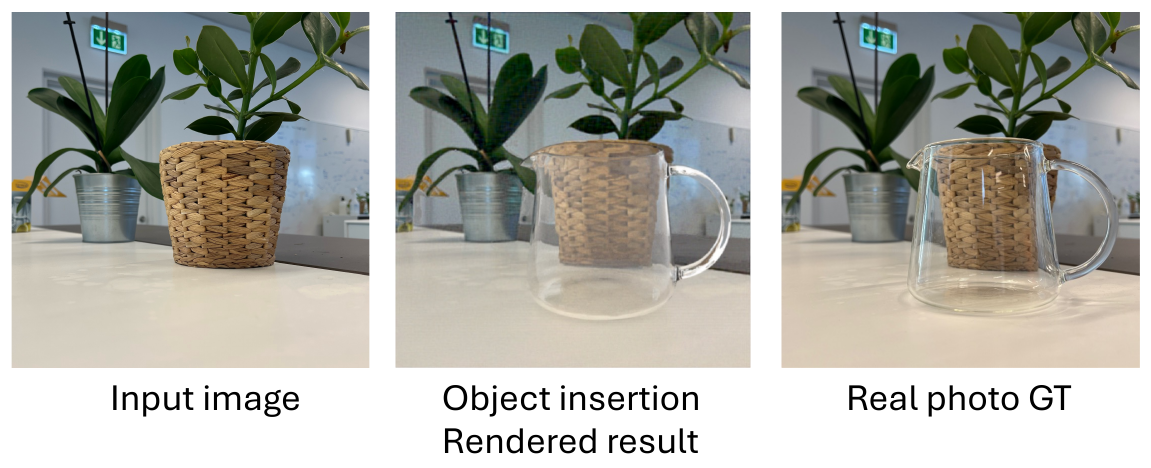}
    \caption{Comparison between GT and our object insertion result for real world image. The error in the reflective area mainly comes from the insufficient accuracy of optimized envmap, as explained in Sec.~\ref{sec:exp_env_opt}.}
    \label{fig:oi_gt}
\end{figure}

Additionally, to validate the visual accuracy of our approach, we used real photograph comparisons instead of synthetic data. We obtained the mesh of a 3D scanned glass jug from previous work~\citep{stets2017scene}, positioned the real jug on a table, and took photographs of the scene from the same angle with and without the jug. We then inserted the mesh jug into the image without the jug and compared it with the actual photograph, as demonstrated in Fig.~\ref{fig:oi_gt}.

Since single-view inverse rendering mostly uses neural networks for rendering and does not support object insertion of transparent objects, we only compare the difference between ground truth and our method.

\begin{figure*}[!tbh]
    \centering
    \includegraphics[width=0.8\linewidth]{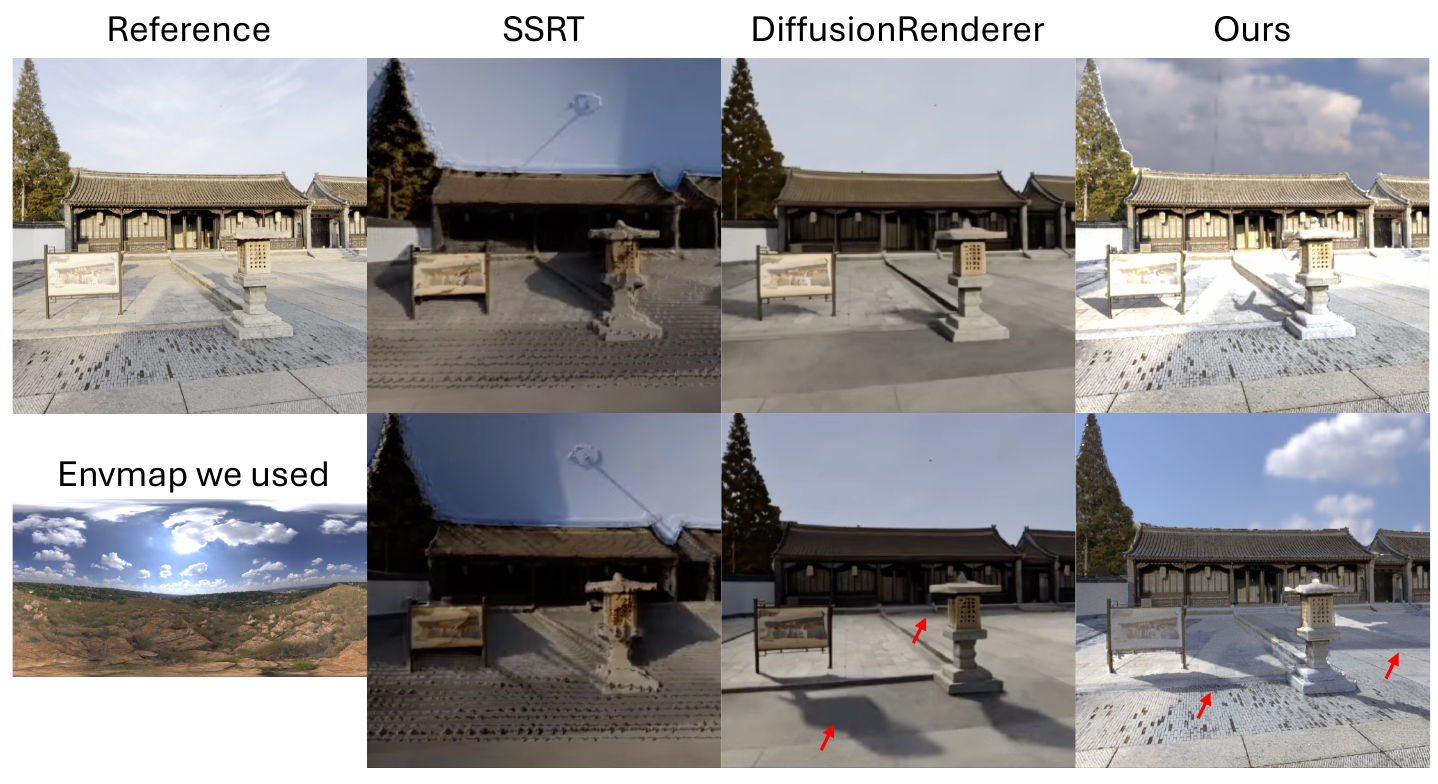}
    \caption{We compare shadow effects with DiffusionRenderer \citep{liang2025diffusionrenderer}. Their images were obtained from their project page. SSRT denotes screen space ray tracing. Note that since the estimated envmaps differ between methods, this comparison highlights geometric consistency rather than absolute lighting accuracy. Specifically, our method preserves the proportional relationship between the shadows of the house and the billboard (red arrow), whereas the baseline produces inconsistent shadow lengths due to the lack of explicit geometry.
    Complex lighting scenarios, such as the soft shadows in the bottom right corner of the image, demonstrate the limitation of our method: because the occluder is outside the view, and there is no occlusion in front of the reconstructed mesh, the shadow is baked into the albedo during optimization. Consequently, even if the envmap is changed, this soft shadow persists.}
    \label{fig:diffusionrenderer}
\end{figure*}

\subsection{Shadow Effects}

Recent advancements in diffusion model-based neural rendering, exemplified by the state-of-the-art work DiffusionRenderer \citep{liang2025diffusionrenderer}, have demonstrated significant improvements in shadow quality. However, due to its reliance on a geometry-free approach, our method, which leverages 3D geometry, continues to achieve superior shadow accuracy, as shown in Fig.~\ref{fig:diffusionrenderer}.

\subsection{Physical Simulation}

\begin{figure}[!tb]
    \centering
    \includegraphics[width=\linewidth]{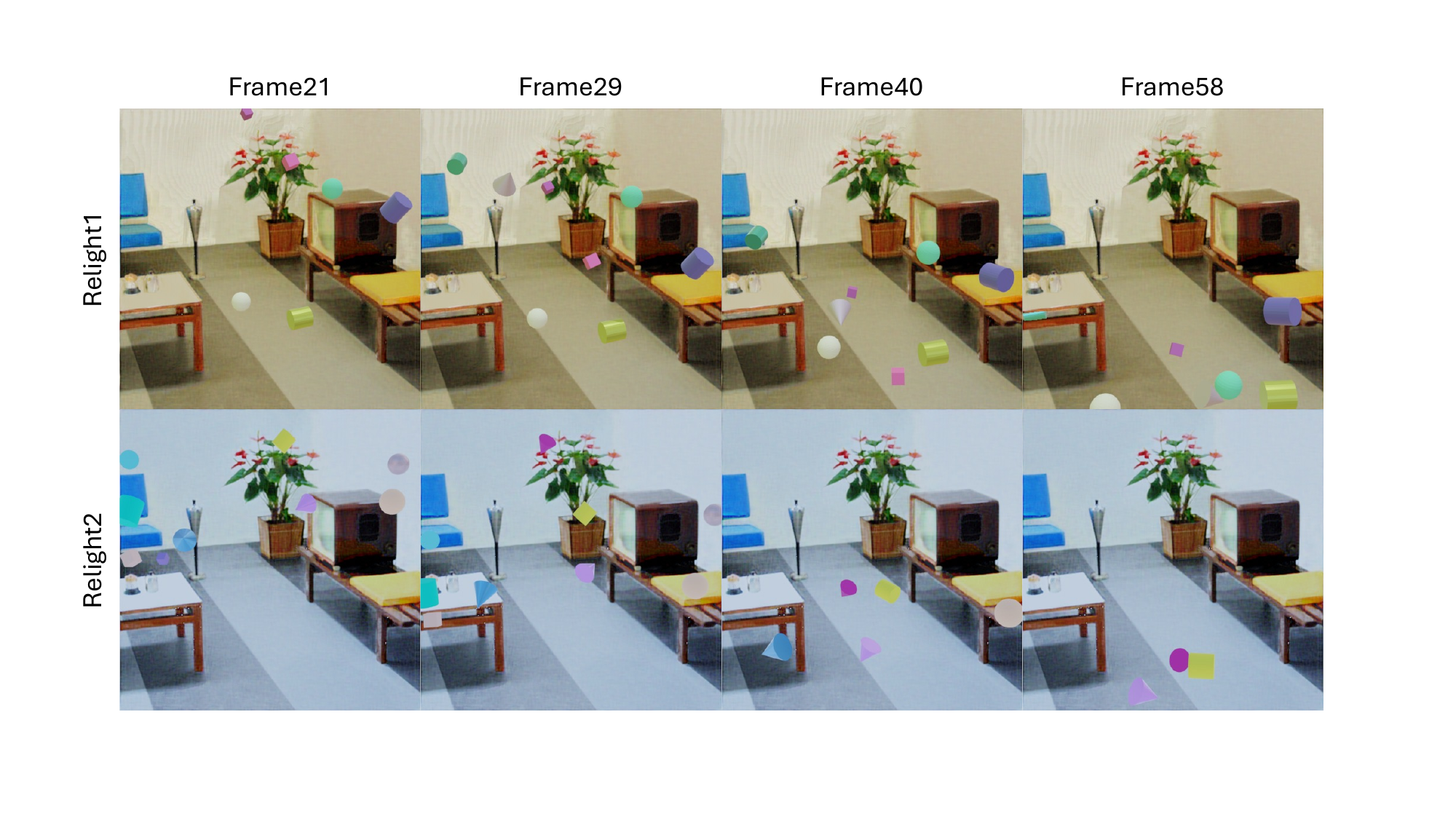}
    \caption{Our methods enable physical simulation. Leveraging 3D geometry, our method enables unique abilities like physical simulation using blender. Note the purple cylinder dropped on the bench and then falls to the floor.}
    \label{fig:phys_sim}
\end{figure}

In contrast to neural rendering approaches \citep{liang2025diffusionrenderer}, our method, benefiting from complete geometry, enables unique physically based simulation capabilities. Fig.~\ref{fig:phys_sim} illustrates a simulation of virtual objects falling in varying illumination conditions, reconstructed from a single image inverse rendering using our methods.

\begin{figure}[!tb]
    \centering
    \includegraphics[width=0.9\linewidth]{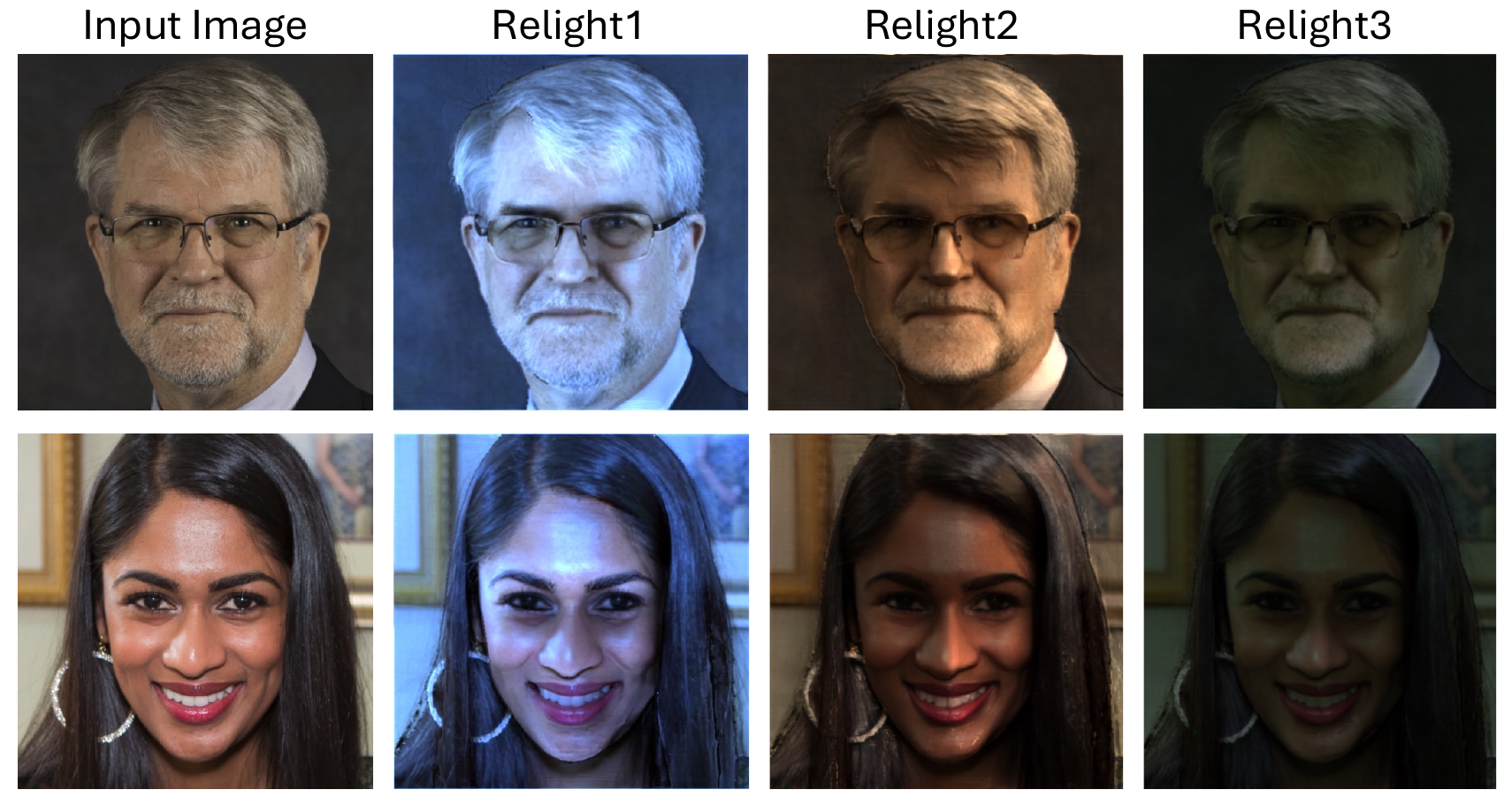}
    
    \caption{Our method was tested on relighting real human faces (FFHQ). When faces lack strong highlights, relighting results are more accurate (above). However, when highlights are present (below), they become baked into the albedo during optimization, causing them to remain under different lighting.}
    \label{fig:face}
    
\end{figure}

\subsection{Human Face Relighting}
We evaluated our method on human face relighting to showcase its potential. Using our MatNet, which is trained on an indoor dataset and not fine-tuned for faces, we predicted $\mathbf{A},\mathbf{R},\mathbf{M}$ and performed progressive optimization for relighting. While results appear promising (Fig.~\ref{fig:face}), our approach does not model subsurface scattering, which is crucial for realistic face rendering. In our method, subsurface scattering is fixed within the albedo during optimization and remains static under lighting changes, limiting accuracy. Similarly, specular highlights on the face are also fixed in the albedo. Future work could improve realism by integrating subsurface scattering into the rendering pipeline.

\begin{figure}[!tb]
    \centering
    
    \includegraphics[width=0.9\linewidth]{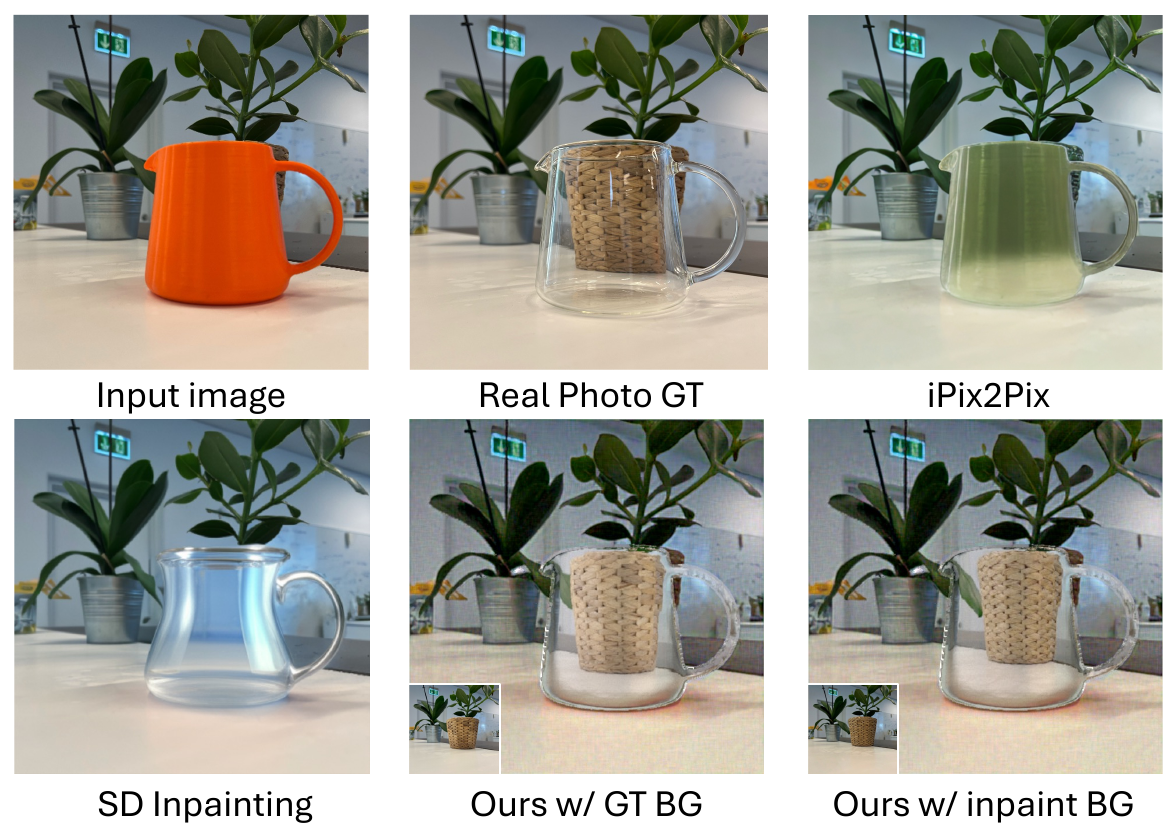}
    \caption{Material transparency editing compared with iPix2Pix \citep{brooks2023instructpix2pix} and DM inpainting \citep{rombach2022high}. Transmission parameter of the DisneyBSDF is set to 0.9, IOR is set to 1.2, and the object color is defined as white. Due to the challenge of obtaining GT background in practical applications, we present results under two scenarios: (1) when the GT background is provided, and (2) when an inpainting-based background is used. As demonstrated, even with the inpainting-based background, our method achieves competitive performance in refraction editing. The bottom-right corner of the image shows the BG used for transparency editing. 
    }
    \label{fig:trans_edit_bm}
\end{figure}

\begin{figure*}[!tb]
    \centering
    \includegraphics[width=0.9\linewidth]{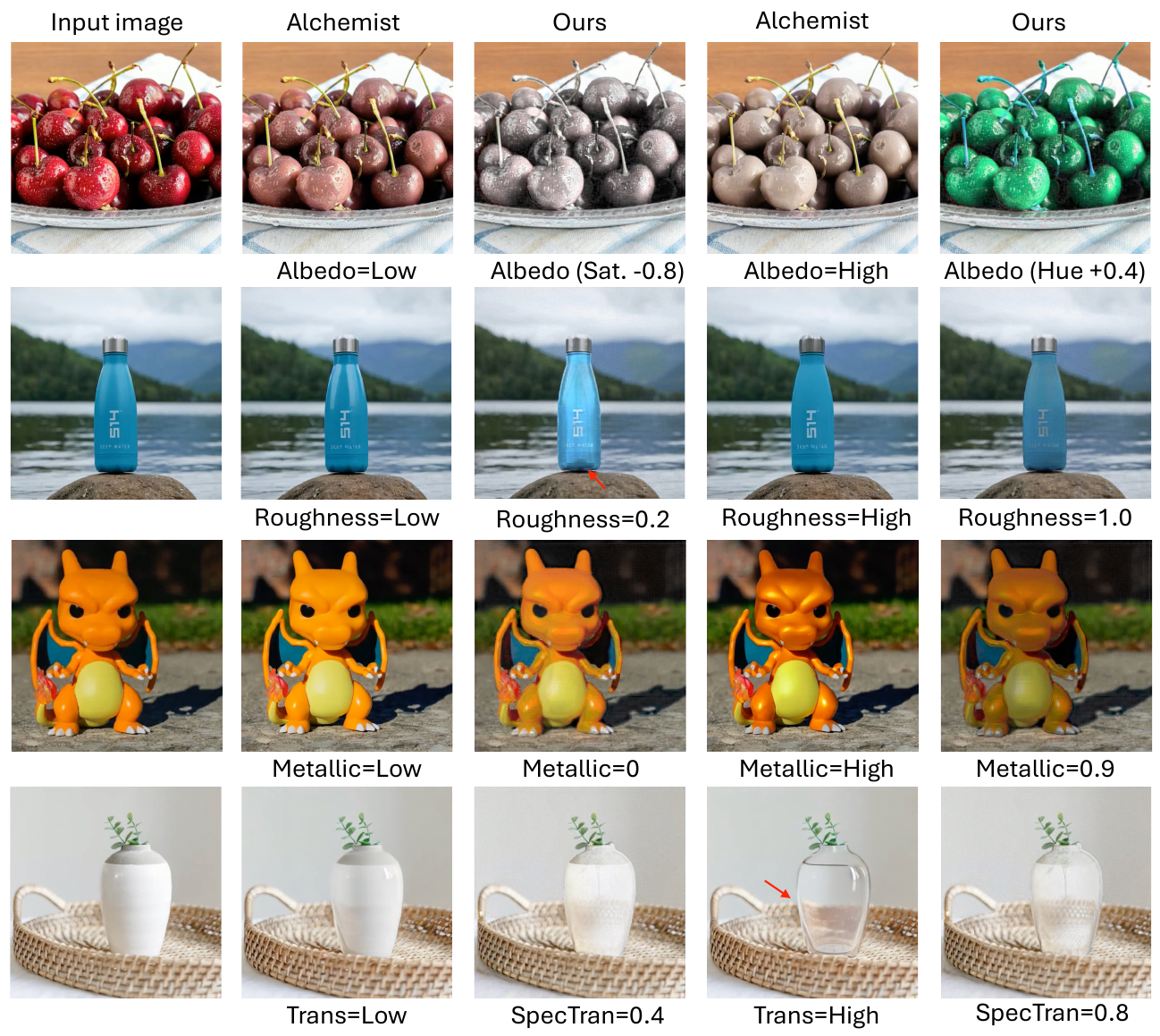}
    \caption{Qualitative evaluation of material editing. 
    Note that as Alchemist~\citep{sharma2024alchemist} is not open-sourced, their results are sourced from official samples, where ``Low'' and ``high'' labels indicate the magnitude of editing. 
    For a fair comparison, our method uses scalar values of $-0.5$ and $+0.5$ to represent equivalent editing intensities for ``Low'' and ``High'' cases, respectively. Our method enables albedo editing to various colors, whereas Alchemist~\citep{sharma2024alchemist} is limited to white. In the roughness editing task, our method correctly renders the reflection of the surrounding environment (e.g., the stone reflected on the bottle, red arrow), ensuring scene integration. Alchemist generates a high-frequency glossy texture but fails to reflect the correct scene context. Our method preserves Charizard's color during metallic edits (third row) and shows that increasing metallic alone darkens the material. In transparency editing, our approach captures light refraction and maintains the vase's appearance when transparency is increased.
   Note that while Alchemist may produce sharper local textures due to its generative nature, it lacks the global physical constraints enforced by our inverse rendering pipeline.}
    \label{fig:mat_edit_bm_al}
\end{figure*}

\subsection{Material Editing} \label{sec:exp_mat_edit}

For transparency editing evaluation, we used real-world photo comparisons. A 3D-printed opaque jug, identical in size to the scanned mesh from Sec.~\ref{sec:oi}, was placed alongside a glass jug at a fixed camera position, and two photos were captured as ground truth for transparency editing.
Since Alchemist ~\citep{sharma2024alchemist}, closely related to our work, is not open-sourced, we compared our method with iPix2Pix~\citep{brooks2023instructpix2pix} and DM inpainting~\citep{rombach2022high}. For iPix2Pix, text prompts were used for editing, while DM inpainting received both a mask of the jug region and a text prompt. Results are in Fig.~\ref{fig:trans_edit_bm}.

For albedo, roughness, and metallic evaluation, 
since the code for Alchemist~\citep{sharma2024alchemist} is unavailable, we utilize the official results provided by the authors for comparison. In their terminology, ``Low'' and ``High'' represent the direction and magnitude of the edit. To maintain consistency, we map these qualitative labels to quantitative scalar inputs in our pipeline: we use $-0.5$ to represent ``Low'' (reduction of property) and $+0.5$ to represent ``High'' (enhancement), providing a perceptually aligned comparison.
 While MatNet predicts accurately for indoor scenes, only albedo predictions remain reliable, with roughness and metallic predictions being less consistent. However, differential rendering allows us to perform inverse rendering on real photos using only albedo.

A note on initialization for editing tasks: for the specific purpose of comparing material editing capabilities, we manually initialized roughness and metallic values to 0.5. This was done to avoid parameter saturation (values converging to 0 or 1), which would otherwise prevent the demonstration of bidirectional editing (increasing or decreasing values). For all other tasks, a default automatic setting is used.
Results are shown in Fig.~\ref{fig:mat_edit_bm_al}.

To ensure a fair comparison, we evaluated our direct rendering results against Alchemist \citep{sharma2024alchemist} without applying super-resolution (SR) post-processing to either method. The results are summarized in Table \ref{tab:edit_user}.
As observed, Alchemist achieves a slightly higher overall score (7.68 vs. 7.36). This is expected, as diffusion-based models excel at generating high-frequency surface details (e.g., complex metallic grains) that are difficult to recover via optimization from a single view. However, the margin is narrow, demonstrating that our physically based approach yields competitive visual realism.
Importantly, while Alchemist produces sharp textures, it often lacks physical consistency. For instance, in roughness editing, our method correctly simulates the reflection of the environment on the object surface (Fig.~\ref{fig:mat_edit_bm_al}, red arrow), whereas diffusion models often generate a generic texture that ignores scene geometry and lighting direction. Similarly, for transparency, our method's reliance on estimated geometry allows for consistent refractive distortions, whereas generative baselines may hallucinate inconsistent background warping. Thus, our method offers a vital alternative for applications requiring physical editability over pure texture hallucination.

\begin{table}[!tb]
    \centering
    \captionsetup{width=\linewidth} 
    \caption{Material editing user study. We compare Alchemist \citep{sharma2024alchemist} against our direct optimization output. Scores indicate user preference (higher is better). Despite lacking a generative prior for high-frequency textures, our method achieves comparable perceptual quality to the diffusion-based baseline.}
    \begin{tabular}{c|c|c}
    \toprule
        Edit Task & Alchemist & Ours (Direct)  \\ \hline
        \rule{0pt}{2.5ex}Albedo & 1.04 & \textbf{2.36}   \\ 
        Roughness & \textbf{2.21} & 1.57   \\ 
        Metallic & \textbf{2.50} & 1.79  \\ 
        Trans & \textbf{1.93} & 1.64  \\  \hline
        \rule{0pt}{2.5ex}Sum & \textbf{7.68} & 7.36  \\ 
    \bottomrule
    \end{tabular}
    \label{tab:edit_user}
\end{table}

\section{Ablations}

\subsection{Evaluation of Transparency Editing}
\label{sec:trans_edit_ab}

To rigorously evaluate the performance of our transparency editing module in a non-privileged setting, we constructed a comprehensive synthetic benchmark. This dataset comprises 20 distinct object geometries, categorized into two groups: 

1. Symmetric Objects: Geometries with front-back symmetry (e.g., spheres, simple vases) where our physical assumption (Front Normal $\approx$ Back Normal) holds well.

2. Asymmetric Objects: Complex geometries (e.g., twisted shapes, statues) with varying thicknesses and concave surfaces, representing challenging cases where physical assumptions are violated.

We rendered these objects as dielectrics (IOR=1.1 and 1.3) under 30 diverse environment maps, yielding a total of 1200 evaluation scenes.
For the input data, we simulated the opaque objects intended for editing by assigning a diffuse material with a uniform RGB albedo of $0.5$. 

\paragraph{Experimental Setup}
To strictly isolate the performance on refractive regions, all metrics (PSNR, SSIM, LPIPS) are computed exclusively within the object mask. We compare two settings to validate the robustness of our pipeline against the lack of ground truth background:
\begin{itemize}
    \item GT Background: Uses the GT background image for refraction calculation. This represents the theoretical upper bound of our rendering module given the geometric estimation.
    \item Inpaint Background: Uses the background generated by the inpainting model \citep{Rombach_2022_CVPR,wu2025qwenimagetechnicalreport}, removing the original opaque object. This reflects the actual use cases.
\end{itemize}

\begin{figure*}[!htbp]
    \centering
    \includegraphics[width=1\linewidth]{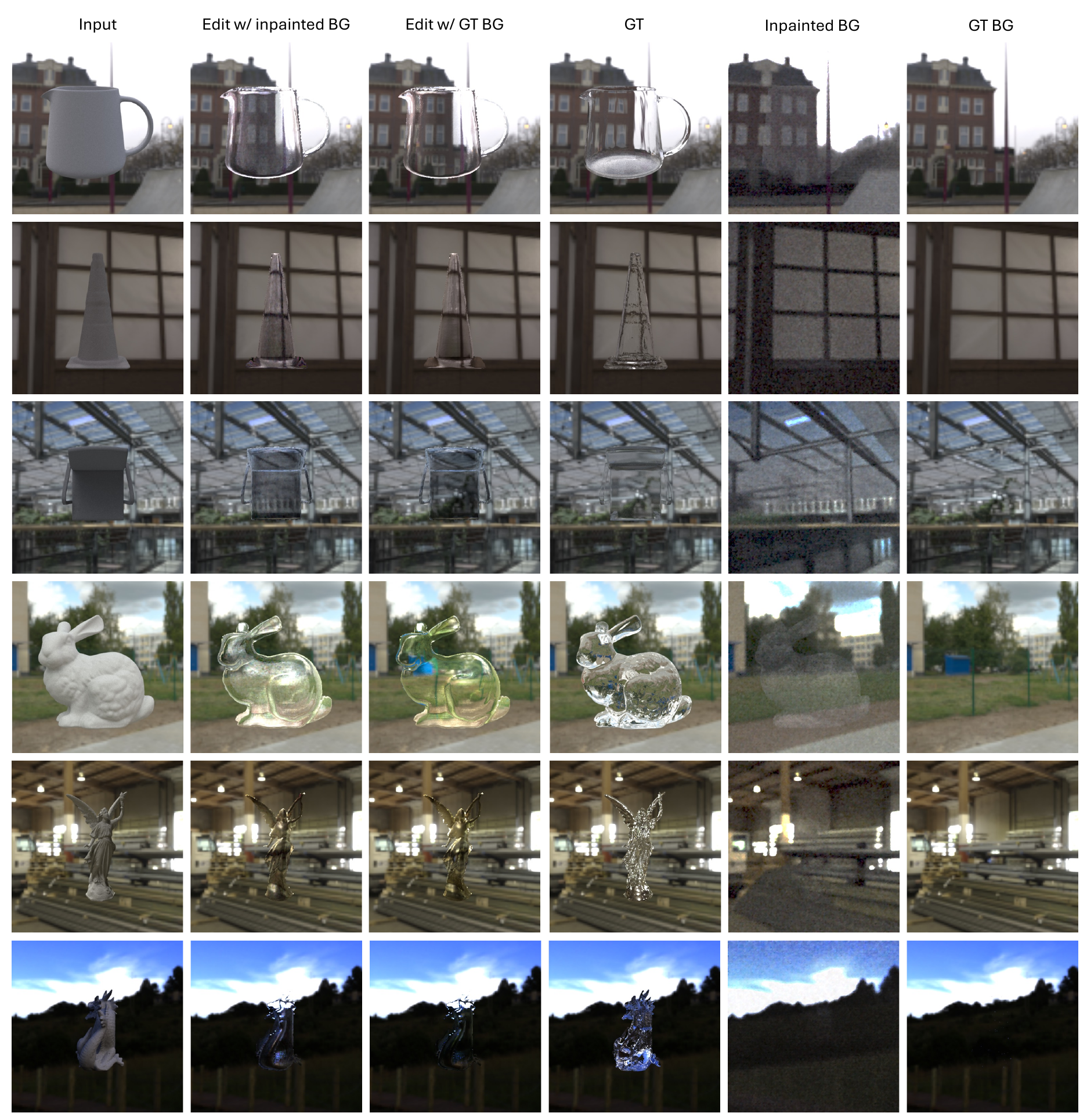}
    \caption{Qualitative results of the transparency editing ablation study. 
    Top three rows (Symmetric Objects): Our method produces transparency effects that closely align with the GT, demonstrating that our simplified refraction model works well for objects with front-back geometric symmetry. 
    Bottom three rows (Asymmetric Objects): The results deviate from the GT. This discrepancy arises because our approach employs a simplified optical model assuming only two refraction events (entry and exit). Consequently, the method fails to account for multiple internal reflections and complex light transport within intricate geometries, which typically accumulate radiance and result in the brighter, ``whitened'' appearance observed in the GT. 
    Effect of background source (4th Row): Comparing the result using ``Inpainted BG'' vs. ``GT BG'', we observe that for complex geometries, the error is dominated by the geometric simplification rather than the background quality. In some specific cases (e.g., 4th row), artifacts from the inpainting process (e.g., residual contours) may coincidentally mimic the complex scattering noise, but quantitatively, both background sources suffer from the limitation of the single-view geometric assumption.}
    \label{fig:trans_blation}
\end{figure*}

\begin{figure}
    \centering
    \includegraphics[width=\linewidth]{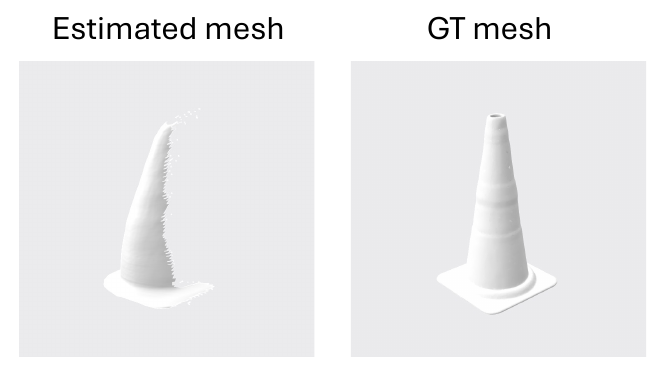}
    \caption{Subtle geometric discrepancies between the estimated mesh and the GT mesh (e.g., slight protrusions on the GT surface) can also lead to inaccuracies in the refraction effects.}
    \label{fig:trans_mesh}
\end{figure}

\begin{table}[!tbp]
  \centering
  \setlength{\tabcolsep}{1mm}
  \captionsetup{width=\linewidth} 
  \caption{Quantitative ablation on background sources and geometric complexity. Metrics are computed on refractive regions only. The small performance gap between GT and Inpaint backgrounds demonstrates the robustness of our method.}
  \label{tab:trans_blation}
  \begin{tabular}{ccccc}
  
    \toprule
    BG SRC & Geometry & PSNR $\uparrow$ & SSIM $\uparrow$ & LPIPS $\downarrow$ \\
    \midrule
    Inpaint & Asymmetric  & 19.278 & 0.739 & 0.316 \\
    Inpaint & Symmetric  & 20.162 & 0.761 & 0.313 \\
    \midrule
    GT   & Asymmetric & 18.667 & 0.751 & 0.314 \\
    GT   & Symmetric & \textbf{20.308} & \textbf{0.793} & \textbf{0.303} \\
    \bottomrule
  \end{tabular}
\end{table}

\paragraph{Results and Analysis}
Table~\ref{tab:trans_blation} presents the quantitative results. We observe two key trends:

1. Robustness to Background Quality.
Comparing the ``Inpainted Background'' and ``GT Background'' settings, the performance degradation is minimal for symmetric objects (e.g., PSNR $20.308$ vs. $20.162$). This confirms that our proposed pipeline is not heavily reliant on privileged background information. The generated backgrounds provided by recent inpainting models are sufficiently plausible to generate convincing refractive effects.

2. Impact of Geometric Assumptions (Failure Case Analysis).
There is a notable performance gap between symmetric and asymmetric objects. Symmetric objects consistently achieve higher fidelity. For asymmetric objects, the performance drops (PSNR $\approx$ 18.667), and notably, using the GT background does not yield better scores than the inpainted background. 
This indicates that for complex geometries with varying thicknesses, the dominant source of error is not the background content, but the violation of the single-view geometric assumption. Our method approximates refraction using a two-interface model based on the front normal. Complex asymmetric shapes induce multiple internal reflections and total internal reflection (TIR), phenomena that accumulate radiance and create bright, ``whitened" appearances in the GT (as seen in Fig.~\ref{fig:trans_blation}). Our simplified model cannot capture these complex light transport paths, leading to darker predictions. Consequently, providing a perfect GT background does not rectify the errors caused by the geometric approximation, making the inpainted background performing comparably.

Our method excels in preserving plausible refractive distortions for general objects using only a single image, though it is fundamentally limited by the single-view geometric ambiguity in highly complex, non-convex shapes.

\subsection{Sensitivity to Camera Intrinsics}
\label{sec:ablation_fov}

Our pipeline assumes a fixed FOV of $35^\circ$, as estimating exact camera intrinsics from a single image is an ill-posed problem. To evaluate the robustness of this assumption, we conducted a stress test by varying the FOV by $\pm 20\%$ (ranging from $28^\circ$ to $42^\circ$) while keeping other hyperparameters constant.

In a single-view setup, there exists an inherent ambiguity between the focal length $f$ and the object depth $Z$. The projection is defined as $u \propto f \cdot \frac{X}{Z}$. An error in the assumed $f$ leads the optimization to converge to a scaled version of the depth $Z$ to maintain alignment with the input image pixels (screen-space supervision). Consequently, while the absolute geometry may be distorted (flattened or elongated along the optical axis), the reprojection remains consistent with the input image.
Because our primary goal is \textit{image editing from a fixed viewpoint} rather than metric 3D reconstruction, this coupled distortion between FOV and Depth cancels out during rendering, preserving the visual fidelity. 

We generated 10 randomized scenes, each scene comprises a horizontal ground plane populated with randomly positioned objects, utilizing a diffuse BSDF with a uniform gray albedo (RGB: 0.5). We rendered these scenes with a FOV of $35^{\circ}$ under 60 distinct environment maps used as GT. Subsequently, mesh reconstruction was performed using varying FOV settings ($28^{\circ}$, $35^{\circ}$, and $42^{\circ}$). The reconstructed meshes were re-rendered using the original envmaps, and the resulting images were compared against the $\text{FOV}=35^{\circ}$ GT to compute quantitative metrics. The results are summarized in Table \ref{tab:fov_syn_ab}. 

\begin{figure}[tb!]
    \centering
    \includegraphics[width=\linewidth]{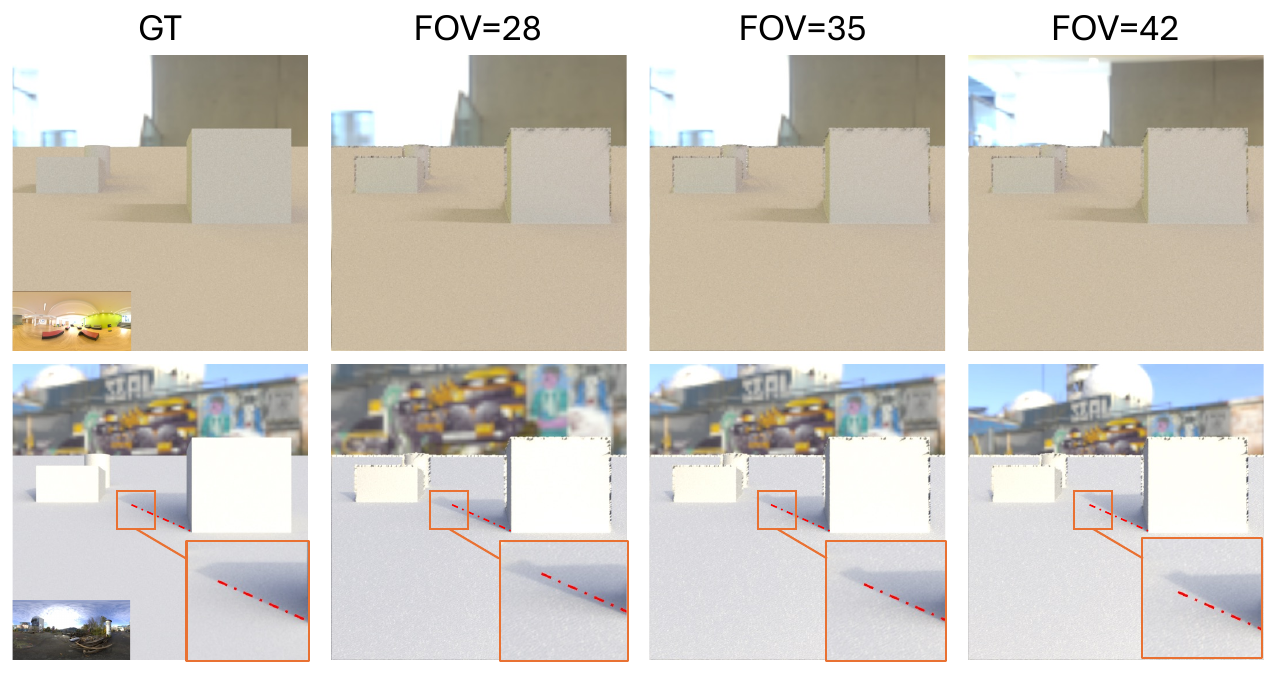}
    \caption{Sensitivity analysis on FOV of synthetic dataset. The meshes reconstructed with different FOVs exhibit almost indistinguishable appearances under different lighting conditions. The red dashed lines in the images in the second row are included to facilitate comparison of the subtle differences in shadows across different FOVs.
}
    \label{fig:fov_ablation_syn}
\end{figure}

As shown in Table \ref{tab:fov_syn_ab}, the quantitative results remain highly consistent across different settings. Specifically, the fluctuation in PSNR is minimal (less than 0.8 dB difference between the extreme cases), and structural metrics (SSIM and LPIPS) show no significant degradation when the initialized FOV deviates from the ground truth. This suggests that our reconstruction pipeline is robust to FOV settings and does not strictly require precise prior knowledge of camera intrinsics. Qualitative result examples are shown in Fig.~\ref{fig:fov_ablation_syn}.

\begin{table}[t]
    \centering
    \captionsetup{width=\linewidth} 
    \caption{Quantitative comparison on synthetic dataset of reconstruction quality under different FOV settings compared against full geometry GT. The results indicate that our method is robust to various FOVs.}
    \label{tab:fov_syn_ab}
    \begin{tabular}{lccc}
        \toprule
        FOV & PSNR $\uparrow$ & SSIM $\uparrow$ & LPIPS $\downarrow$ \\
        \midrule
        28$^{\circ}$ & 28.50 & 0.758 & 0.203 \\
        35$^{\circ}$  & 28.56 & 0.759 & 0.202 \\
        42$^{\circ}$ & 27.85 & 0.751 & 0.213 \\
        \bottomrule
    \end{tabular}
\end{table}

\begin{figure*}[tbh!]
    \centering
    \includegraphics[width=\linewidth]{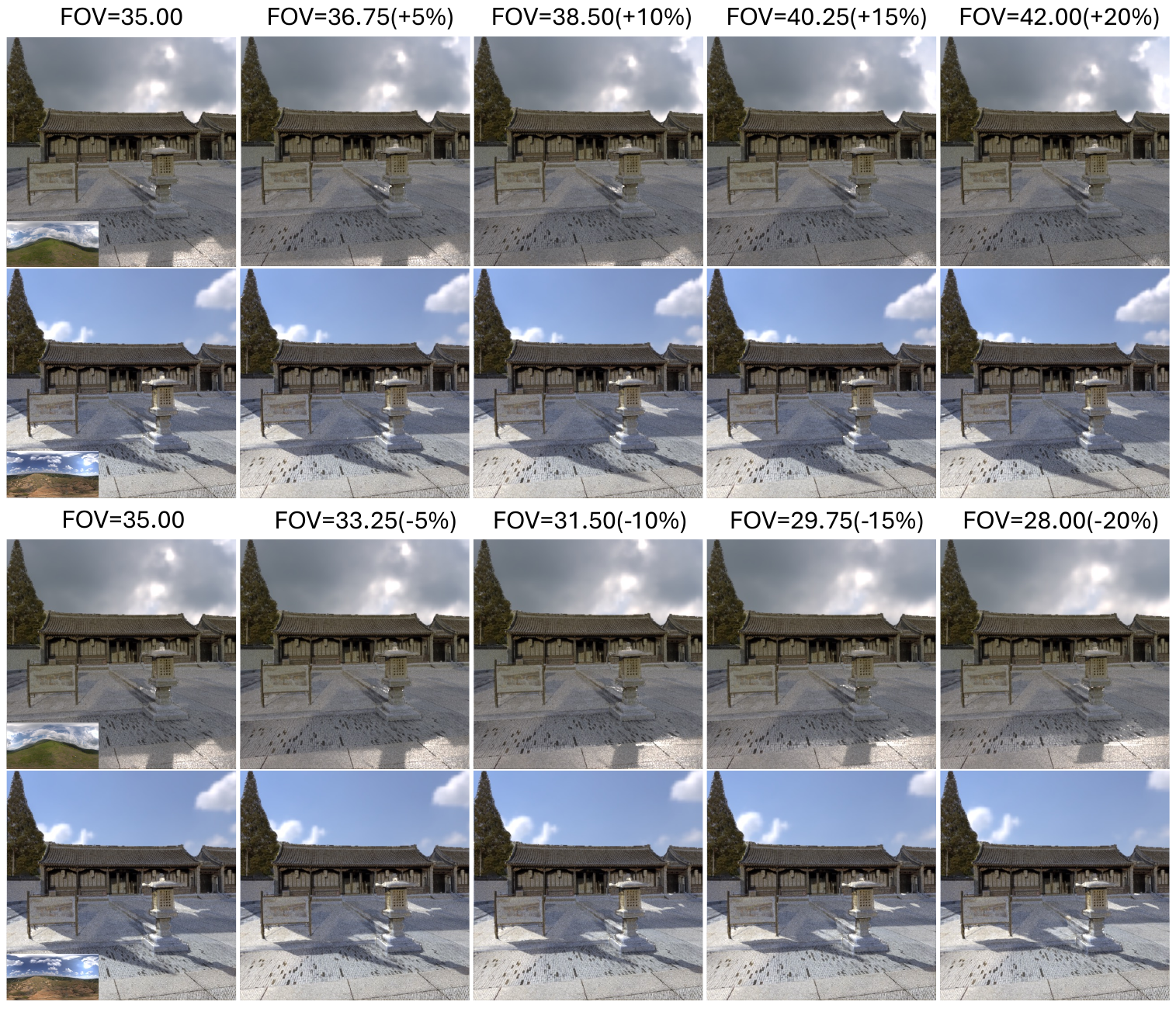}
    \caption{Sensitivity Analysis of FOV on real world image. The envmap used is shown in the lower-left corner of the first column, and the same envmap is applied to all images in each row. We vary the assumed FOV by $\pm 20\%$ ($28^\circ$, $35^\circ$, $42^\circ$). The first and second rows illustrate FOV increasing from 35° to 42°, while the third and fourth rows demonstrate FOV decreasing from 35° to 28°. The first and third rows depict cases where shadows in the relit results are not prominent, resulting in minimal differences across all images. The second and fourth rows present results where environment maps generate distinct shadows, revealing differences in shadow characteristics. Specifically, a narrower FOV (telephoto-like) produces slightly more parallel shadows, while a wider FOV exhibits greater perspective divergence. Additionally, the difference in FOV can be observed in the sky's cloud distribution, with smaller FOVs capturing less clouds compared to larger FOVs.}
    \label{fig:fov_ablation}
\end{figure*}

We also compare the effects of different FOVs on real-world image relighting.
Fig.~\ref{fig:fov_ablation} illustrates the relighting results under different FOVs. When shadows are absent or not prominent in the images, the differences between images captured with varying FOV settings become almost imperceptible. This is because our boundary-handling mesh reconstruction ensures that texture mapping remains pixel-aligned regardless of the global depth scaling.
On the other hand, varying the FOV alters the perspective projection of cast shadows. A wider FOV results in more divergent shadows, while a narrower FOV produces more parallel shadows. However, since the ``distorted'' shadows are physically consistent with the ``distorted'' geometry induced by the wrong FOV, the resulting edits remain perceptually plausible to the human eye, which lacks a reference for the absolute scene scale. 

This analysis confirms that for the task of image editing from a fixed viewpoint, our method is robust to reasonable deviations in camera intrinsics.

\begin{table*}[!htb]
    \centering
    \caption{Ablation study on envmap settings. 
    We compare Cartesian (UV) vs.\ spherical coordinates across Low ($32 \times 16$) and High ($512 \times 256$) resolutions.
    While Low Res models achieve better numerical fit on the envmap itself (due to smoothing), the High Res Spherical model achieves the best relighting quality.}
    \setlength{\tabcolsep}{1.2mm}
    \begin{tabular}{l|c|cc|cc|cc|c}
    \toprule
        ~ & ~ & \multicolumn{2}{c|}{SSIM $\uparrow$} & \multicolumn{2}{c|}{PSNR $\uparrow$} & \multicolumn{2}{c|}{MSE $\downarrow$} & SH Error $\downarrow$ \\ 
        Setting & Res & Envmap & Render & Envmap & Render & Envmap & Render & Envmap \\ \hline
        Ours (UV) & Low & 0.516 & 0.810 & 17.023 & 19.888 & 0.178 & 0.039 & 0.570 \\
        Ours (Spherical) & Low & \textbf{0.628} & \textbf{0.811} & \textbf{17.230} & 19.956 & \textbf{0.144} & 0.034 & 0.520 \\ \hline
        Ours (UV) & High & 0.504 & 0.808 & 17.070 & 19.872 & 0.866 & 0.038 & 0.503 \\
        \textbf{Ours (Spherical)} & \textbf{High} & 0.520 & \textbf{0.811} & 17.175 & \textbf{20.048} & 0.791 & \textbf{0.029} & \textbf{0.474} \\
    \bottomrule
    \end{tabular}
    \label{tab:ablation_env_settings}
\end{table*}

\subsection{Envmap Optimization} \label{sec:ablation_env}

We investigate the impact of coordinate system parameterization and envmap resolution on the optimization performance. Where no specific instructions are provided, the experimental settings are maintained in alignment with the configuration outlined in Section \ref{sec:exp_env_opt}. The results are summarized in Table~\ref{tab:ablation_env_settings}.

\paragraph{Coordinate Systems} 
We compare the performance of using standard 2D Cartesian (UV) coordinates versus the proposed spherical 3D unit vectors for positional encoding. 
Comparing \textit{Ours (UV)} and \textit{Ours (Spherical)} at the same high resolution ($512 \times 256$), the spherical parameterization consistently yields lower SH Error ($0.472$ vs $0.503$) and higher Rerender PSNR ($20.05$ vs $19.87$). 
This indicates that mapping inputs to a continuous spherical domain provides a better inductive bias for learning omnidirectional lighting, effectively eliminating discontinuities at image boundaries and singularities at the poles. The qualitative results are presented in Fig.~\ref{fig:env_bm}.

\paragraph{Resolution}
We evaluate the performance of different envmap resolution.
At a low resolution ($32 \times 16$), the optimization is strictly constrained to low-frequency signals. Consequently, it achieves a very low Envmap MSE ($0.144$) and high Envmap SSIM ($0.628$) by fitting a smooth mean of the lighting.
However, this smoothness comes at the cost of missing details required for realistic reflections.
By increasing the resolution to $512 \times 256$, although the pixel-wise Envmap MSE increases due to the increased high-frequency details, the Rerender PSNR improves to $20.05$.

\paragraph{Predicted Material Properties on Envmap Estimation}
We tested the results of envmap optimization under two conditions: providing MatNet predictions versus setting $\mathbf{A}=\mathbf{R}=0.5, \mathbf{M}=0.1$. As shown in Table~\ref{tab:envmap_bm}, providing MatNet predictions significantly improves the accuracy of envmap estimation.

\subsection{Scaling Factor $\delta$}

\begin{table}[!tb]
    \centering
    \caption{Ablation of $\delta$ in Eq.~\eqref{eq:cons_loss}. Comparing the rerender error. Optimization steps are limited to 200.}
    \setlength{\tabcolsep}{1mm}    
    \begin{tabular}{c|ccc|ccc}

    \toprule
        ~ & \multicolumn{3}{c|}{Opt $\mathbf{ARM}$} & \multicolumn{3}{c}{Opt $\mathbf{RM}$\&$\mathbf{A}$} \\ 
        $\delta$ & SSIM $\uparrow$ & PSNR $\uparrow$ & MSE $\downarrow$ & SSIM $\uparrow$ & PSNR $\uparrow$ & MSE $\downarrow$ \\ \hline
        0.0 & 0.459 & 13.479 & 0.0504 & 0.611 & 17.390 & 0.0342 \\ 
        0.5 & 0.473 & 14.490 & 0.0473 & \textbf{0.699} & \textbf{18.242} & \textbf{0.0271}  \\ 
        1.0 & 0.542 & 15.363 & 0.0426 & 0.618 & 17.396 & 0.0312  \\ 
        1.5 & 0.554 & 15.403 & 0.0446 & 0.637 & 17.541 & 0.0313  \\ 
        2.0 & 0.571 & 15.320 & 0.0424 & 0.640 & 17.719 & 0.0280  \\ 
        2.5 & 0.610 & \textbf{16.151} & 0.0371 & 0.640 & 16.922 & 0.0304  \\ 
        3.0 & \textbf{0.612} & 16.109 & \textbf{0.0366} & 0.656 & 16.819 & 0.0299 \\ 
    \bottomrule
    \end{tabular}
    \label{tab:delta_ab}
\end{table}
We conducted an ablation study on the value of $\delta$ in Eq.~\eqref{eq:cons_loss}. Using 20 randomly selected images from the IIW dataset~\citep{bell14intrinsic}, we tested different $\delta$ values and compared the rerendered images with the input images. We evaluated two strategies: simultaneous optimization of $\mathbf{ARM}$ and sequential optimization ($\mathbf{RM\&A}$), where $\mathbf{RM}$ is optimized first, followed by $\mathbf{A}$. To save time, we applied early stopping, halting optimization if $\mathcal{L}_{\text{re}}$ reduced by less than 5\% over 20 consecutive steps. Results are shown in Table~\ref{tab:delta_ab}. 

\subsection{Weights Initialization} \label{sec:ab_weights}
\begin{figure}
    \centering
    \includegraphics[width=\linewidth]{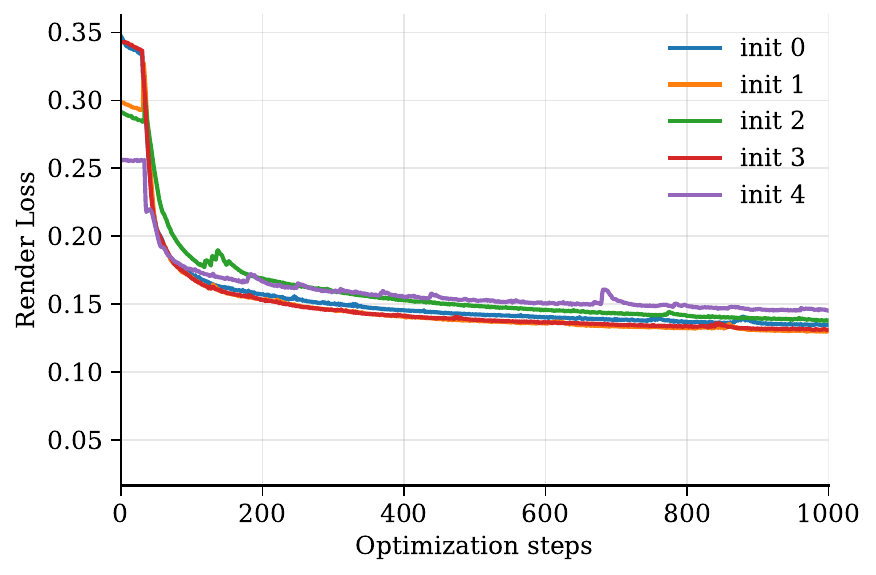}
    \caption{Ablation of different weight initialization. The lines in different colors represent distinct initial values. The effect of varying neural network initializations on the optimization process is limited.}
    \label{fig:ab_init}
\end{figure}

We perform a comprehensive ablation study to investigate the impact of various weight initialization strategies for the optimization neural network (MLP), while maintaining consistent material parameter initializations derived from MatNet predictions. We tested five distinct random seeds for initializing the weights of different neural networks. As shown in Fig.~\ref{fig:ab_init}, the optimization process's loss demonstrates that variations in neural network configurations have limited influence on the optimization outcome. This is anticipated, as the neural network is solely employed to accelerate the optimization process.

\subsection{Geometric Quality} \label{sec:geo_qual}

\begin{figure}
    \centering
    \includegraphics[width=\linewidth]{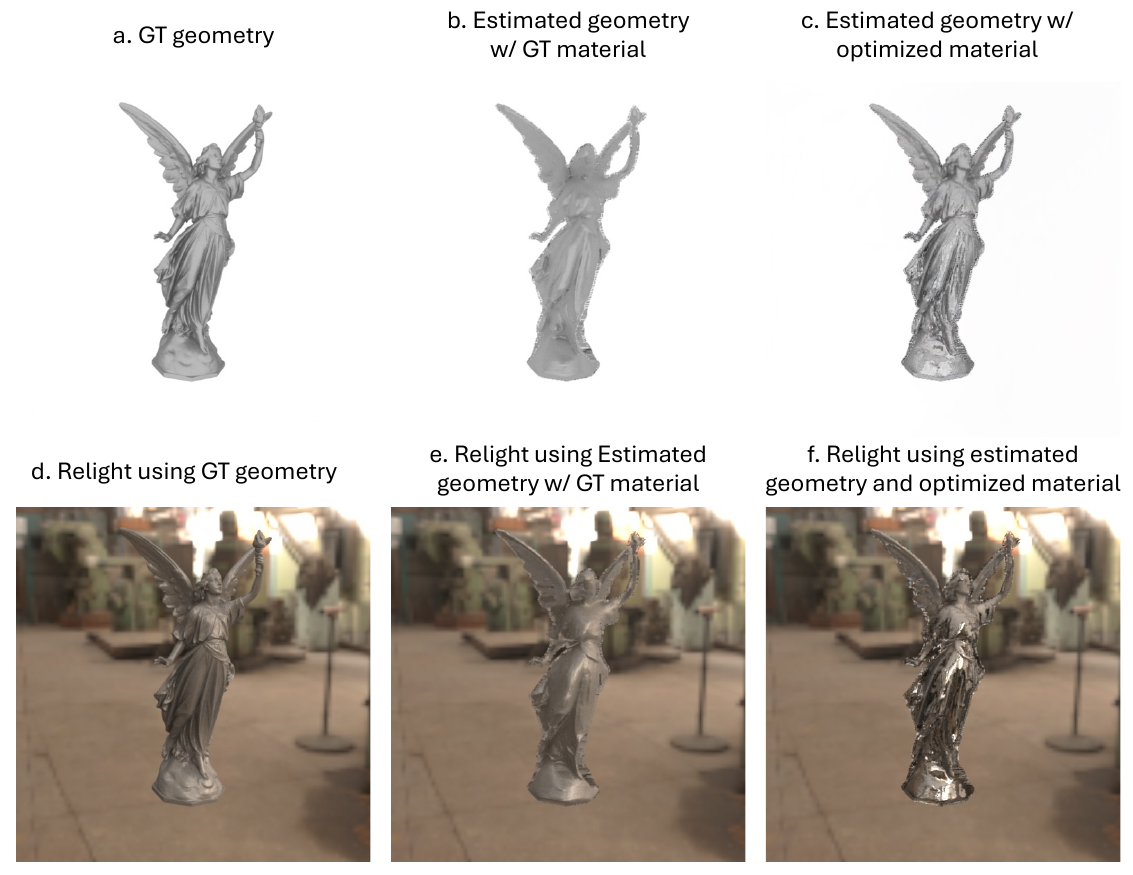}
    \caption{Qualitative ablation of geometric quality. (a) The ground truth reference. (b) Rendering using the estimated geometry with GT material properties, showing significant deviation. (c) Our final result after material optimization, which closely matches the GT in the original lighting. However, compensating for geometric inaccuracies via material properties (e.g., lower roughness for highlights) leads to visible artifacts under a new lighting condition (f).}
    \label{fig:geo_qual}
\end{figure}

Our method's reliance on geometry allows it to outperform the concurrent work, DiffusionRenderer \citep{liang2025diffusionrenderer}, particularly in the rendering of shadow effects. However, this dependency introduces an associated limitation. We conduct an ablation study to analyze the impact of geometric quality. To clearly illustrate this effect, we utilize a more challenging geometric configuration for inverse rendering with our proposed method, as depicted in Fig. \ref{fig:geo_qual}.

Fig. \ref{fig:geo_qual}(b) displays the estimated geometry. When rendered with GT material properties, the result exhibits significant deviations from the GT. In contrast, Fig. \ref{fig:geo_qual}(c) presents the result after optimizing the material properties using the estimated geometry, which more closely approximates the GT. Due to the lack of fine details in the estimated geometry, the material properties must be optimized to compensate for these geometric inaccuracies. For instance, the highlights on the skirt in Fig. \ref{fig:geo_qual}(c) are achieved by reducing the material's roughness rather than by accurate geometric representation. However, this compensation comes at a cost. When the optimized scene is subsequently relighted, noticeable artifacts emerge, as seen in the relit example in Fig.~\ref{fig:geo_qual}(f).

Fig.~\ref{fig:geo_qual} highlights a fundamental trade-off in our method: the material optimization process compensates for inaccuracies in the initial geometry prediction. While this compensation successfully minimizes the reconstruction error under the original lighting, the resulting material maps may not be physically accurate and can lead to artifacts when the scene is relit. Thus, the quality of our relighting and editing results is closely tied to the geometric accuracy provided by MatNet. Improving single-view geometry estimation would directly enhance the robustness of our material optimization.

\subsection{SH Coefficients for Illumination Representation}

We tested using SH coefficients instead of envmaps and found that optimizing SH coefficients took longer to reach the same loss. We use 3rd-order SH coefficients to represent the light source. Given the ground truth material properties, we perform light source optimization under different lighting conditions. The time taken by each method to achieve the same $ \mathcal{L}_{\text{re}}$ is compared. The results is shown in Table \ref{tab:sh}. 

\begin{table}[!htbp]
    \centering
    \captionsetup{width=\linewidth} 
    \caption{Time taken by different methods to optimize  $\mathcal{L}_{\text{re}}$ to 0.03. Direct envmap optimization is faster than optimize SH coefficient.}
    \begin{tabular}{c|cc}
        \toprule
         ~ & SH coeff & Envmap  \\ \hline
         Time & 473 s & 103 s \\
         \bottomrule
    \end{tabular}
    \label{tab:sh}
\end{table}

\subsection{Optimization Steps and Rerendering Error}

We analyzed the relationship between optimization steps and re-render error. 
We randomly selected 10 images from the InteriorVerse dataset \citep{zhu2022learning} to test re-rendering error at different optimization steps. Optimization on an RTX 3090 runs at 2 steps per second. Based on MatNet's accuracy, satisfactory results are typically achieved after 10 minutes of optimization.
Results are shown in Table~\ref{tab:steps_ab}.

\begin{table}[!ht]
    \centering
    \captionsetup{width=\linewidth} 
    \caption{Ablation of optimization steps. Metrics are based on re-rendering error.}
    \begin{tabular}{c|ccc}
    \toprule
        Opt Steps & SSIM $\uparrow$ & PSNR $\uparrow$ & LPIPS $\downarrow$ \\ \hline
        50 & 0.645 & 15.529 & 0.180  \\
        200 & 0.789 & 20.124 & 0.134  \\ 
        1000 & 0.885 & 27.675 & 0.109  \\ 
        2000 & 0.943 & 33.204 & 0.093 \\ 
        \bottomrule
    \end{tabular}
    \label{tab:steps_ab}
\end{table}

\subsection{Optimization Network Type}

\begin{figure}[!htbp]
    \centering
    \includegraphics[width=1\linewidth]{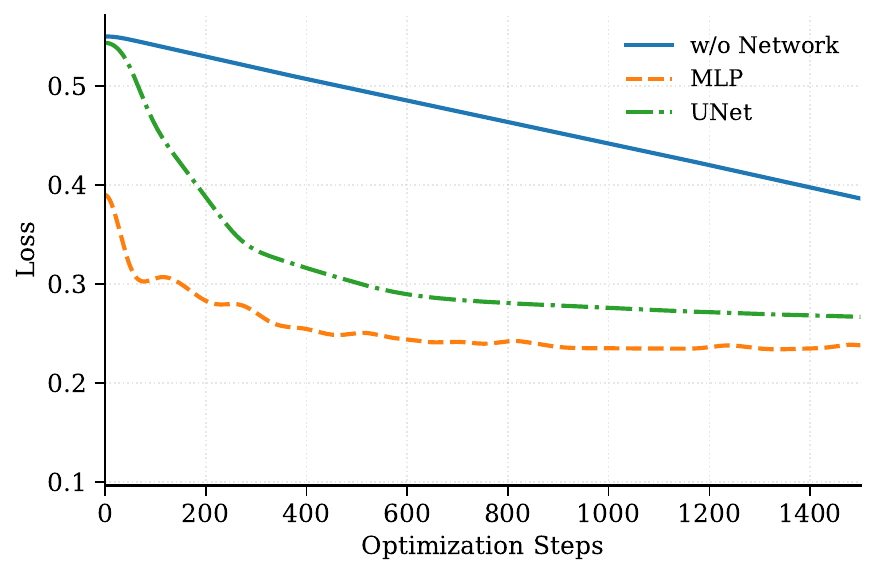}
    \caption{Compared to a CNN-based UNet, the position-embedded MLP achieves faster convergence.}
    \label{fig:net_ab}
\end{figure}

We experimented with using a CNN-based UNet for material properties optimization. The UNet architecture includes 2 downsampling blocks and 2 upsampling blocks, with each block containing two 2D convolution layers (kernel size = 3, padding = 1). Additionally, we compared direct optimization of material properties without a neural network. As shown in Fig. \ref{fig:net_ab}, direct optimization was the slowest, followed by the CNN-based UNet, while the position-embedded MLP achieved the fastest convergence speed. For additional studies, see the supplementary material.

\begin{figure}
    \centering
    \includegraphics[width=1\linewidth]{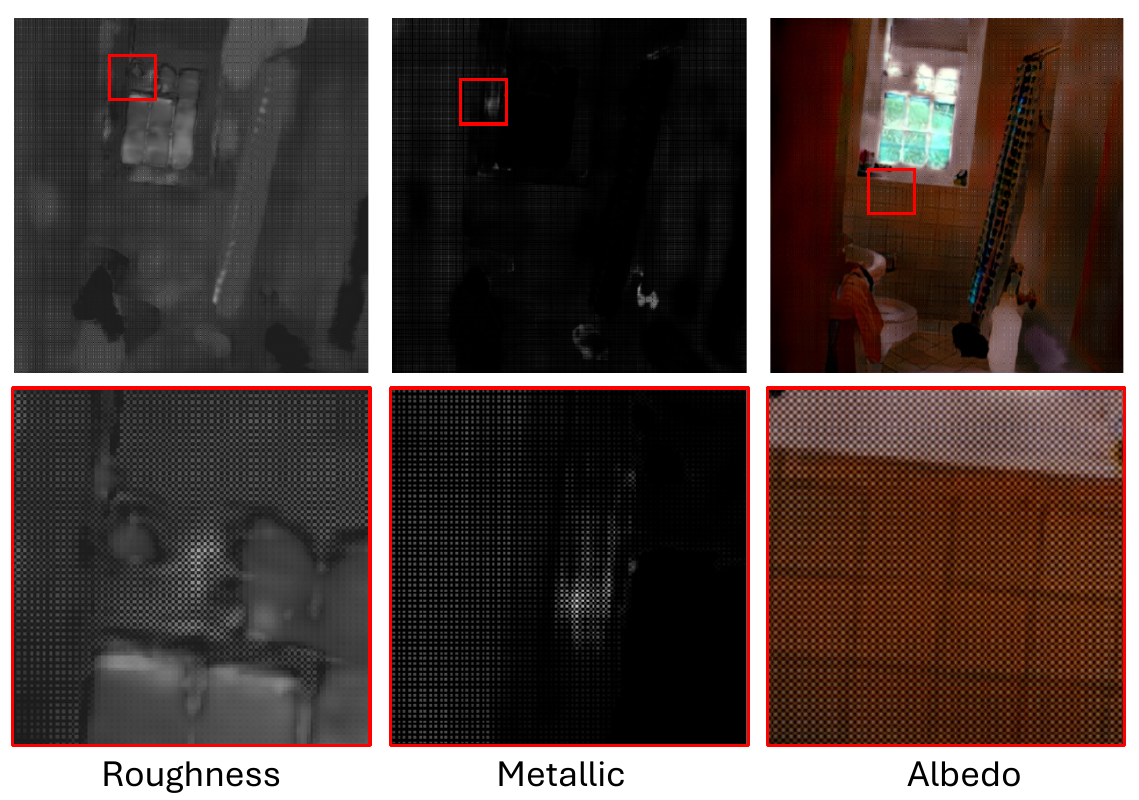}
    \caption{UNet Artifect. The red box highlights an enlarged view of this area, and the details is in the second row.}
    \label{fig:unet_artifect}
\end{figure}

Additionally, due to the characteristics of convolutional neural networks, noticeable artifacts may appear during optimization. As shown in Fig. \ref{fig:unet_artifect}, the optimized images show a regular pattern of black dots. 

It is worth noting that when MatNet predictions are unreliable (e.g., for roughness and metallic), directly optimizing material properties without relying on the neural network can achieve lower $\mathcal{L}_{\text{re}}$  given sufficient optimization time (such as more than 3000 optimization steps). Table \ref{tab:min_L_re} shows the minimum MSE render loss achievable by both methods in the absence of roughness and metallic predictions.

\begin{table}[!ht]
    \centering
    \captionsetup{width=\linewidth} 
    \caption{Comparison of the lowest $\mathcal{L}_\text{re}$ that can be obtained using MLP and directly optimizing material properties when the predicted roughness and metallic are not provided.}
    \begin{tabular}{c|c|c}
    \toprule
         ~ & w/o Network & w/ MLP  \\ \hline
        Min $\mathcal{L}_{\text{re}}$ & 0.0038 & 0.0087 \\
    \bottomrule
    \end{tabular}
    
    \label{tab:min_L_re}
\end{table}

\section{Limitations} \label{sec:limits}

In this section, we discuss potential limitations of our proposed method from a broader, user-oriented perspective. It is important to note that some points, such as the need for per-image optimization, are not typically viewed as limitations within the differentiable rendering field. However, from a user’s viewpoint, this may make our approach less convenient than DM-based methods \citep{sharma2024alchemist,brooks2023instructpix2pix}.

1. As with all differentiable rendering methods, our approach requires optimization for each image to obtain the lighting and optimized material properties, which might be quite time-consuming. Depending on MatNet's prediction accuracy, optimization takes between 5 to 30 minutes. For out-of-domain images (i.e., non-indoor scenes) or higher-quality requirements, optimization time may increase. This limits our method's ability to achieve the rapid, batch image editing enabled by stable diffusion-based techniques \citep{brooks2023instructpix2pix}. Training MatNet on a broader dataset for more accurate predictions could help mitigate this limitation.
 
2. Our method uses an envmap to represent lighting, complicating accurate modeling of physical light sources in the scene. If a light source is present in the original image, optimization may fix it within the albedo, leading to inaccuracies during relighting.

3. The optimization quality of the envmap is highly dependent on the accuracy of material prediction. Since MatNet is trained on a synthetic dataset, when handling out-of-domain real-world images, any bias in material prediction (e.g., misclassifying specular highlights as texture) may lead the envmap optimization to compensate for such errors, thereby resulting in inaccurate estimations. Consequently, performance in real-world scenarios is constrained by the generalization capability of MatNet.

4. The introduction of differentiable rendering allows our method to perform well even on out-of-domain images. However, if MatNet predictions are poor, differential rendering optimization requires careful tuning for each image, such as adding a mask to the object to be edited or experimenting with different optimization strategies.

5. For material transparency editing, complex object geometry can lead to inaccurate refraction distortions, as our method assumes only two refractions. Single-view mesh reconstruction \citep{shi2023zero123plus,liu2023syncdreamer,long2024wonder3d} may yield more accurate refractions for complex shapes. 
Besides, since we simulate refraction without the rays truly passing through the object, shadows cast by transparent objects may lack accuracy, and does not have caustic effect. (Note that this problem does not exist for object insertion tasks with complete geometry.)

6. Our current framework is designed to edit the material of opaque objects, including making them transparent. However, it does not explicitly handle scenes that already contain transparent objects in the input image. In such cases, complex light transport effects like refraction and caustics are likely to be ``baked" into the initial albedo prediction by MatNet. Decomposing these effects from a single view is a significant, ill-posed challenge. This represents a promising, albeit difficult, direction for future research, potentially requiring new priors or network architectures specifically designed for transparent and specular light transport.

7. Due to mesh reconstruction limitations, artifacts may appear along object edges during strong relighting, as the mesh is discontinuous at these edges. This can be mitigated using super resolution.

8. A key limitation of our method, as with all single-image differentiable rendering approaches, is the inherent ambiguity in disentangling object colors, which can arise from either lighting or material properties. While our optimization process operates under these constraints, we mitigate this ambiguity through the use of priors and techniques (Sec. \ref{sec:opt_set}). However, fully resolving this challenge remains beyond the scope of our work. Despite this limitation, our method achieves superior image editing results.
In an ideal scenario, if MatNet could infer ground truth materials directly, optimization would be unnecessary. Nonetheless, our approach strikes a practical balance between performance and feasibility under these inherent constraints.

\begin{figure}
    \centering
    \includegraphics[width=0.9\linewidth]{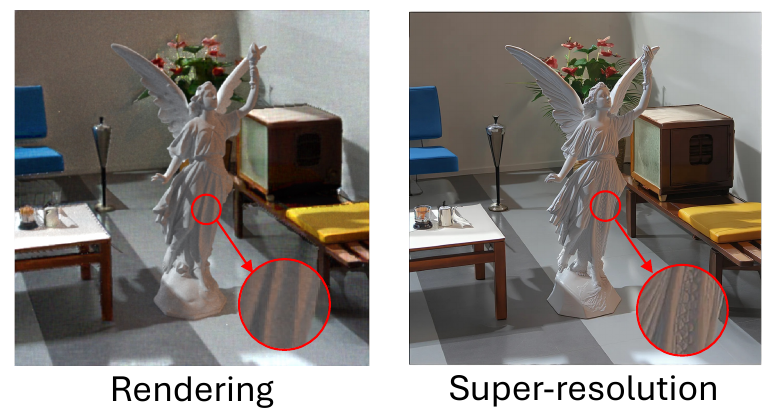}
    \caption{Super-resolution can improve image quality while preserving overall structure, such as lighting and shadows. However, DM-based super-resolution may alter some details, as seen in the excessive modifications to the statue in the right image. In the encircled region, the diffusion model has introduced excessive details that were not present in the original image. Here, we see non-existent textures on the hem of the skirt.
    }
    \label{fig:sr}
\end{figure}

9. Since our images are rendered and mesh reconstruction limitations can cause artifacts along object edges, using SR \citep{wang2021real,rombach2022high,flux2024,wu2024unique3d} may help reduce these artifacts. However, SR models often overly modify the image, as shown in Fig.~\ref{fig:sr}. Improving SR models by giving additional input could potentially mitigate this issue.

10. In some cases, our method can cause double shadows. Even when the predicted albedo contains no shadow, under certain conditions (as demonstrated in Figure \ref{fig:diffusionrenderer}, where there are no occluder in the scene), the shadow can still be baked into the albedo during optimization.
This can lead to ``double shadows" during relighting. 

\section{Conclusion}
We presented a novel method for inverse rendering that combines neural network predictions with differentiable rendering to optimize material properties and lighting conditions from a single image. Our approach enables more accurate envmap estimation, scene level relighting, object insertion, and material editing including transparency editing without requiring complete geometry. % Comparative evaluations demonstrate that our method achieves high fidelity in various tasks, offering a more physical accurate solution for single-view material editing.

\section*{Acknowledgements}
This project has received funding from the European Union’s Horizon 2020 research and innovation programme under the Marie Skłodowska-Curie grant agreement No. 956585 (PRIME). \\
This project has received funding from Innovation Fund Denmark, project 0223-00041B (ExCheQuER). \\
We thank the anonymous reviewers for their
constructive comments, which have significantly
improved the quality of this paper.

\bibliography{references} % common bib file

\clearpage
\begin{appendices}

\section{More Ablations}

\subsection{Impact of Envmap Resolution on Highly Specular Materials} \label{sec:envmap_specular}

\begin{figure}[tbh!]
    \centering
    \includegraphics[width=\linewidth]{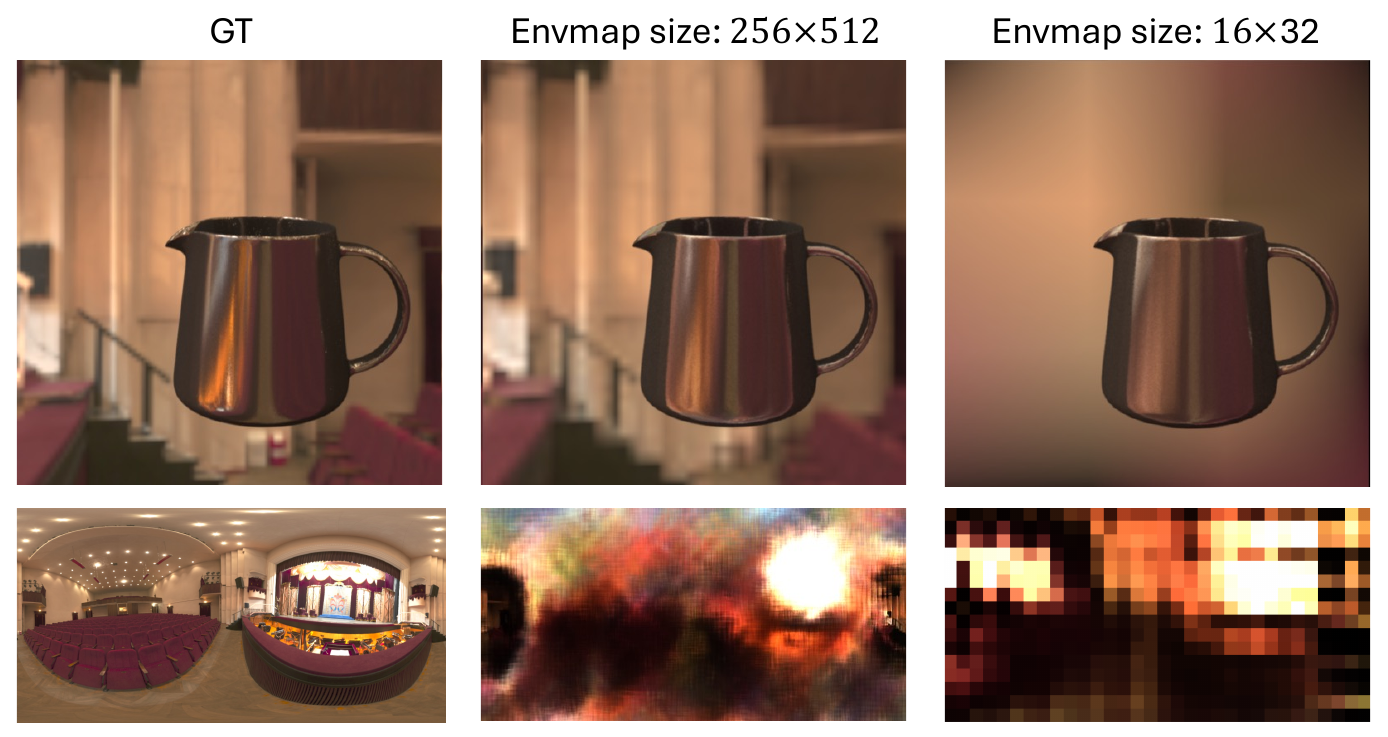}
    \caption{Qualitative ablation of envmap resolution on high reflective surface. Employing a higher-resolution envmap yields rendering results that more closely approximate the ground truth, particularly evident in the orange area to the left of the jug in the image.}
    \label{fig:envmap_size}
\end{figure}

While lower-resolution envmaps typically exert a limited influence on rendering results, this is not the case for highly reflective surfaces. We conducted an ablation study to assess the impact of varying envmap resolutions on such surfaces. Importantly, we maintained ground truth material properties and optimized solely with respect to the envmap. The results is shown in Fig. \ref{fig:envmap_size} and Table \ref{tab:ab_envmap}.

\begin{table}[!ht]
    \centering
    \captionsetup{width=\linewidth} 
    \caption{Quantitative ablation of envmap resolution on high reflective surface. The metrics reported between the rendered results and the GT excluded the background from the evaluation.}
    \begin{tabular}{c|ccc}
    \toprule
        Envmap Size & SSIM $\uparrow$ & PSNR $\uparrow$ & LPIPS $\downarrow$ \\ \hline
        $32\times16$ & 0.911 & 25.042 & 0.143  \\ 
        $512 \times 256$ & 0.967 & 26.540 & 0.074  \\
        \bottomrule
    \end{tabular}
    \label{tab:ab_envmap}
\end{table}

\subsection{Robustness of Mesh Reconstruction} 
\label{sec:bd_ablation}

\begin{figure*}[tbh!]
    \centering
    \includegraphics[width=1.0\linewidth]{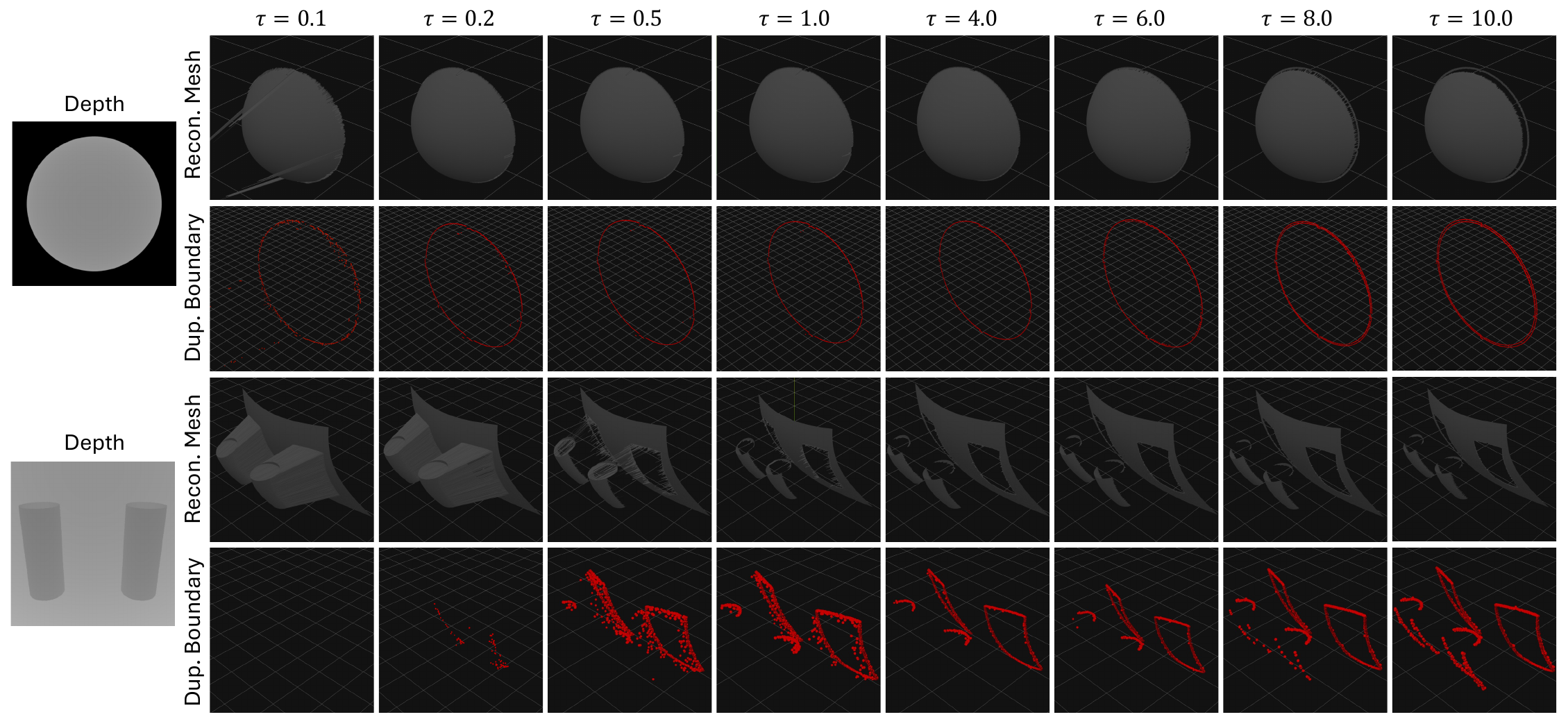}
    \caption{Robustness of Boundary Duplication over $\tau$. 
    (Left) At low $\tau$ ($< 1^\circ$), the threshold is too strict, causing the foreground object to remain topologically connected to the background (``webbing''). 
    (Right) At high $\tau$ ($> 8^\circ$), the condition is too sensitive for high-curvature surfaces. Note the artifacts on the sphere surface, where the curvature is misinterpreted as an occlusion boundary, leading to ``cracking'' or concentric ring artifacts. 
    (Middle) $\tau \approx 6^\circ$ offers an optimal balance for general scenes.}
    \label{fig:tau_ablation}
\end{figure*}

\begin{figure*}[tbh!]
    \centering
    \includegraphics[width=0.8\linewidth]{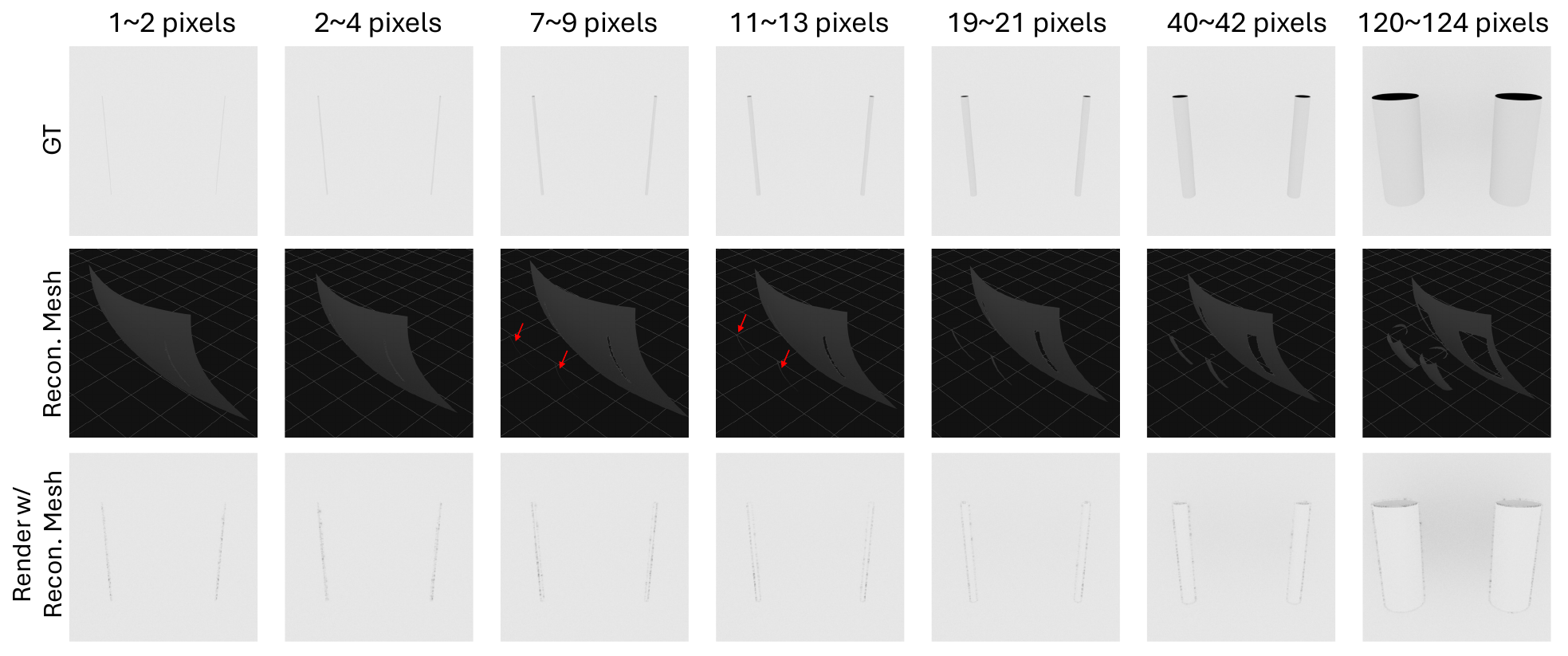}
    \caption{Failure Case: Thin Structures. We evaluate reconstruction stability on cylindrical objects of varying pixel widths. For structures thinner than 7 pixels, the rasterized normals become unreliable, causing the boundary detection to fail. Consequently, the thin object merges with the background surface, resulting in distortions during relighting.}
    \label{fig:bd_thickness}
\end{figure*}

We evaluate the robustness of our Boundary Duplication (BD) method regarding the angular threshold $\tau$ and object thickness.

\paragraph{Sensitivity to $\tau$.} The threshold $\tau$ controls the classification of ``grazing angle" vertices. As shown in Fig. \ref{fig:tau_ablation}, this introduces a trade-off between topological connectivity and geometric fidelity:
\begin{itemize}
    \item Under-segmentation ($\tau < 2^\circ$): An overly strict threshold fails to detect the occlusion boundary at the cylinder's edge, causing the foreground mesh to remain topologically connected to the background (``webbing''). This distorts the texture during relighting.
    \item Over-segmentation ($\tau > 8^\circ$): An overly loose threshold incorrectly classifies the surfaces of high-curvature objects (e.g., the sphere in the top row) as boundaries. This results in ``cracking'' artifacts, where the mesh splits open on continuous surfaces.
    \item Optimal Range: We empirically found that $\tau \approx 6^\circ$ robustly separates foreground objects while preserving the continuity of curved surfaces across our dataset.
\end{itemize}

\paragraph{Failure Case: Thin Structures.} 
We explicitly analyze the resolution limits of our reconstruction in Fig.~\ref{fig:bd_thickness}. Since our boundary detection relies on surface normals derived from the depth map, it depends on the local gradient.
For structures with a screen-space width of less than 7 pixels (Fig.~\ref{fig:bd_thickness}, Right), the discrete gradient estimation becomes unstable. The normals at the object edge are smoothed with the background, causing the boundary condition $\arcsin(\vec{n} \cdot \vec{v}) < \tau$ to fail. Consequently, thin objects may merge with the background surface. This remains a limitation for extremely fine-scale geometry in single-view inverse rendering.

\subsection{Post-Processing with Super-Resolution}

\begin{figure*}
    \centering
    \includegraphics[width=1\linewidth]{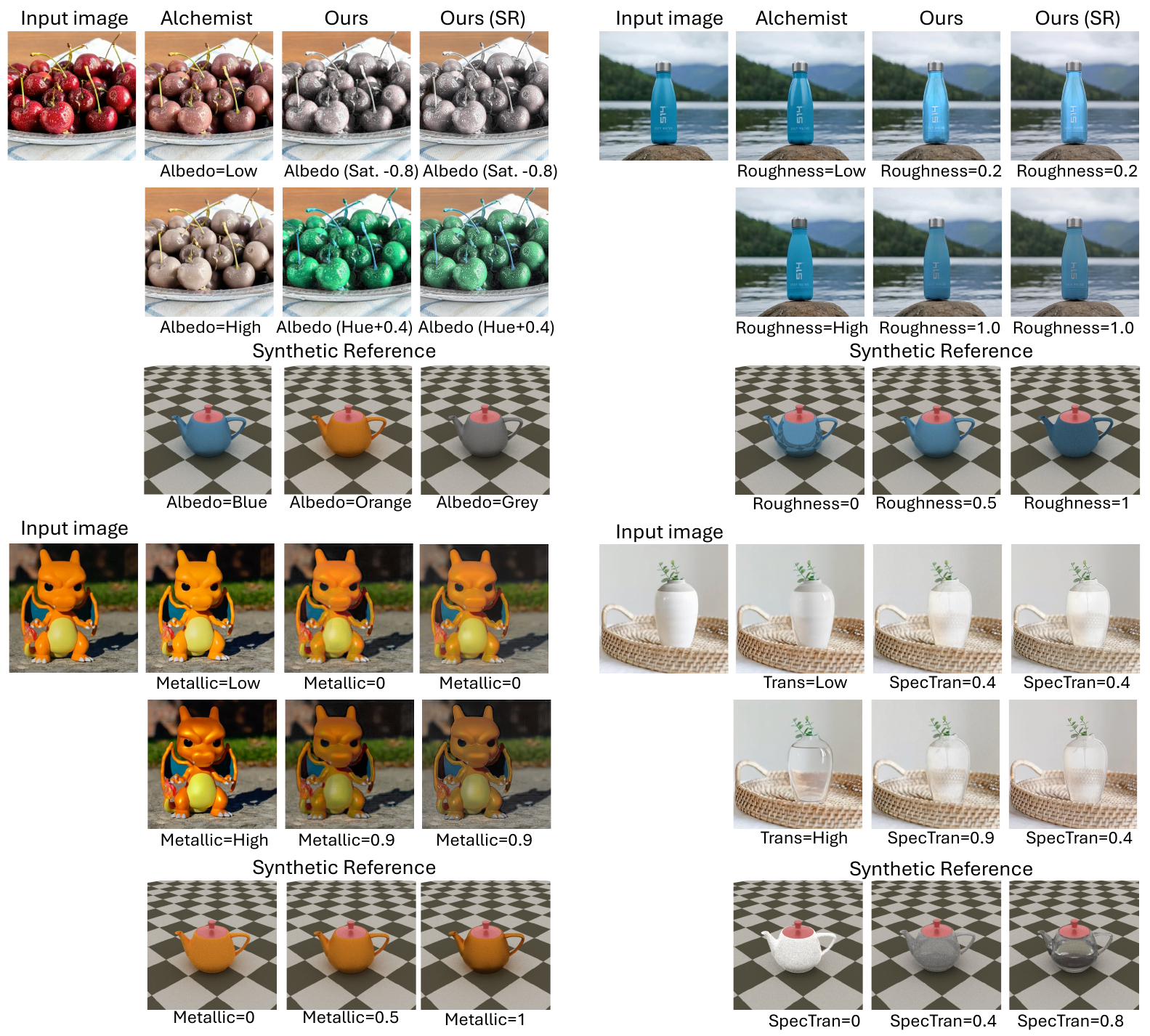}
    \caption{\textbf{Impact of Super-Resolution on Rendered Images.} While super-resolution (SR) can significantly enhance the visual quality of our rendered outputs by smoothing artifacts, current models often fail to accurately preserve original details, introducing unintended modifications. This figure also showcases how deliberate edits to material properties affect an object's final appearance in a synthetic scene, illustrating the trade-off between artifact reduction and content fidelity when applying SR.}
    \label{fig:mat_edit_appendix}
\end{figure*}

Our rendering approach, combined with mesh reconstruction, can produce artifacts along object edges. We explore using super-resolution (SR)~\citep{wang2021real,rombach2022high,flux2024,wu2024unique3d} to mitigate these issues and enhance visual quality. However, as demonstrated in Fig.~\ref{fig:mat_edit_appendix}, standard SR models often struggle to preserve the fidelity of the input, leading to undesirable alterations. We posit that conditioning SR models with additional information, such as material or geometric maps, could alleviate this problem and improve content preservation. 
Our user study (Table~\ref{tab:edit_user_ab}) demonstrates that the application of SR leads to a substantial improvement in user ratings, increasing from 7.36 to 8.96.
\begin{table}[!htb]
    \centering
    \captionsetup{width=\linewidth} 
    \caption{Material Editing User Study. SR represent for super resolution.}
    \begin{tabular}{c|c|c}
    \toprule
        Edit Task &  Baseline & Baseline w/ SR  \\ \hline
        Albedo & 2.36 & \textbf{2.61}  \\ 
        Roughness & 1.57 & \textbf{2.21}  \\ 
        Metallic  & \textbf{1.79} & 1.71  \\ 
        Trans &  1.64 & \textbf{2.43}  \\  \hline
        Sum & 7.36 & \textbf{8.96} \\ 
    \bottomrule
    \end{tabular}
    \label{tab:edit_user_ab}
\end{table}

\section{Coordinate Transformation} \label{sec:screeb_coor}

Since our inputs for the renderer are in screen space, we briefly describe how to transform a world-space position $\bm{x}$ into screen space. 
Extending $\bm{x}$ to homogeneous coordinates $\mathbf{X} = [x, y, z, 1]^T$, we first transform into camera space coordinates using 
\begin{equation}
\mathbf{Y} = \bm{V} \, \mathbf{X} \,,
\end{equation}
where $\bm{V}$ is the view matrix (the homogeneous form of the extrinsic camera matrix).
For a perspective projection matrix $\bm{P}$, we obtain normalized device coordinates (NDC) by
\begin{eqnarray}
\mathbf{Q} & = & \bm{P} \, \mathbf{Y} \\
\mathbf{N}_\text{ndc} & = & \mathbf{Q}/w \,,
\end{eqnarray}
where $w$ is the fourth coordinate of $\mathbf{Q}$. Using the first two components of $\mathbf{N}_\text{ndc}$ ($x_{\text{ndc}}$ and $y_{\text{ndc}}$), we obtain the screen-space coordinates as follows:
\begin{eqnarray}
   x_{\text{screen}} & = & \left( \frac{x_{\text{ndc}} + 1}{2} \right) \cdot W \\
   y_{\text{screen}} & = & \left( \frac{y_{\text{ndc}} + 1}{2} \right) \cdot H \,.
\end{eqnarray}
This process maps a world position to a specific location on the screen, which we denote as $\bm{x}_s = \mathbf{S}(\bm{x}) = [x_{\text{screen}}, y_{\text{screen}}]$. Thus, for a shading point~$\bm{x}$, we use screen coordinate indexing to get the material properties, i.e., $\mathbf{A}(\bm{x}_s),\mathbf{R}(\bm{x}_s)$, $\mathbf{M}(\bm{x}_s)$.

\section{Handling HDR Images}
\label{sec:hdr_handling}

HDR images, characterized by their wide range of luminance levels, pose stability challenges during the differentiable rendering optimizations. The extensive variation in pixel intensities can lead to unstable gradients, making optimization difficult. To mitigate this issue, a common practice is to clip the image intensity values to a predefined range, typically $[0, 10]$. This clipping helps to limit the extreme values that can cause numerical instability during backpropagation.
In cases where the images are extremely bright, additional preprocessing steps are often necessary. One effective approach is to apply a logarithmic transform, specifically $\log(1+x)$, to the image data. This transformation compresses the dynamic range of the image, reducing the disparity between the brightest and darkest regions. By doing so, the gradients become smoother and more manageable, facilitating more stable and efficient optimization. The $\log(1+x)$ transform is particularly useful because it preserves the relative differences in intensity while compressing the overall range, ensuring that the model can learn effectively from the data without being overwhelmed by extreme values.

\section{Refraction Network Details} \label{sec:RefNet}

\subsection{Dataset Preparation}

To train the RefractionNet, we constructed a synthetic dataset using physically based rendering.

The dataset comprises 50 meshes of common transparent objects (e.g., glassware, liquids) rendered from 200 random viewpoints.
For each view, we compute the Ground Truth Optical Path Length $d_\text{GT}$ by tracing rays from the camera. 
The calculation logic distinguishes between solid and hollow objects:
\begin{equation}
    d = 
    \begin{cases}
        \| \mathcal{P}_{\text{exit}} - \mathcal{P}_{\text{entry}} \| , & \text{Solid Object} \\
        & \text{\small(e.g., crystal ball)} \\
        \| \mathcal{P}_{\text{inner\_entry}} - \mathcal{P}_{\text{entry}} \| &\\
        {}+\| \mathcal{P}_{\text{exit}} - \mathcal{P}_{\text{inner\_exit}} \|, & \text{Hollow Object} \\
        & \text{\small(e.g., empty bottle)} \\
    \end{cases}
\end{equation}
where $\mathcal{P}$ denotes intersection points. This allows the network to learn distinct thickness profiles for different object typologies.

\begin{figure}[!tbh]
    \centering
    \includegraphics[width=0.9\linewidth]{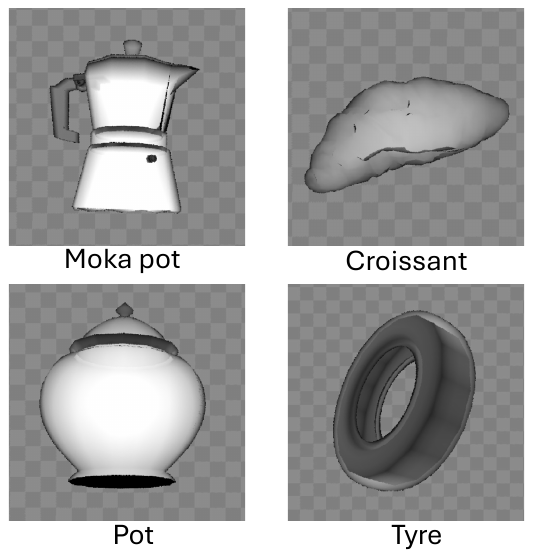}
    \caption{Example of refraction thickness dataset. The closer to white, the longer the refraction distance.}
    \label{fig:refr_path}
\end{figure}

Examples are shown in Fig. \ref{fig:refr_path}.

\begin{figure*}[!tbh]
    \centering
    \includegraphics[width=0.8\linewidth]
    {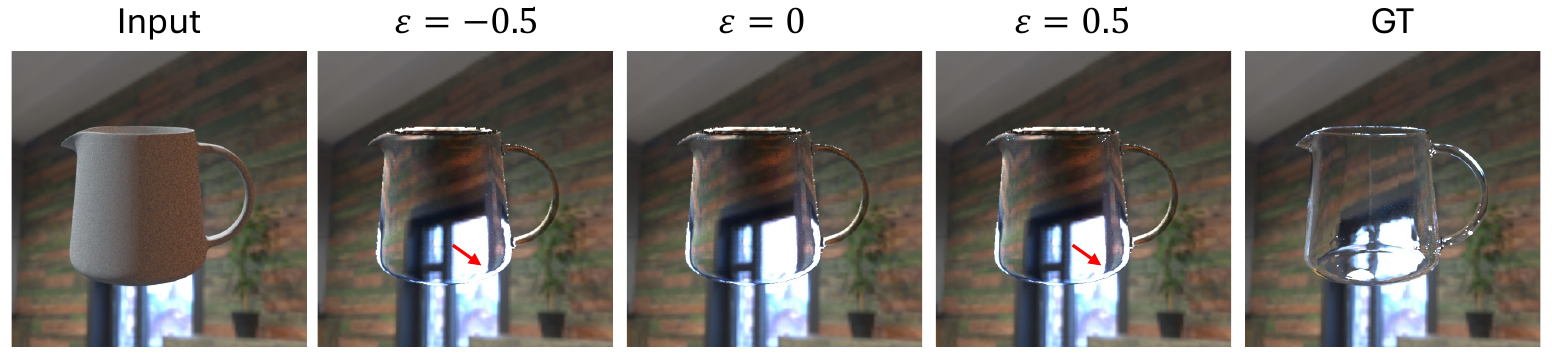}
        \caption{Sensitivity analysis of refraction depth. Errors up to 50\% in thickness prediction produce visually plausible results. The red arrows indicate the subtle differences in the results for varying values of $d$.}
    \label{fig:refr_sens}
\end{figure*}

\subsection{Sensitivity Analysis of $d$}
\label{sec:sensitivity_d}

We conducted a sensitivity analysis of the refraction thickness $d$ using five objects with bilateral symmetry. 
A full geometry mesh is rendered using diffuse materials with an envmap as the light source, serving as the input image for the transparency editing task. The same mesh is set as dielectric materials with an IOR of 1.3 to render the GT for transparency editing.
We establish a baseline by rendering transparency effects using the ``Front Normal $\approx$ Back Normal'' assumption (which our pipeline uses).
To mitigate the influence of inpainted backgrounds on the results, we use GT backgrounds for the transparency editing task. Specifically, we first performed transparency editing by adding noise to d. Subsequently, we introduced random noise to d according to Equation \eqref{eq:ablation_d} and repeated the transparency editing process 10 times. The rendered results were then evaluated and compared against GT to assess the robustness of the refraction thickness parameter: 
\begin{equation}
    \label{eq:ablation_d}
    d_\text{noise} = d_\text{pred} \cdot (1 + \varepsilon), \quad \varepsilon \in [-0.5, 0.5] \,.
\end{equation}

As shown in Fig. \ref{fig:refr_sens}, errors in $d$ shift the sampling location of the background texture (indicated by red arrows) with minimal distance. Human perception of refraction is primarily driven by high-frequency distortions and specular highlights. The highlights depend on the front surface (which we recover accurately from MatNet). Since the background in practical editing scenarios is inpainted (and thus lacks a ground truth reference for the user), these shifts are perceptually interpreted as natural refractive distortions rather than errors. This confirms that a learned prior for $d$ is sufficient for convincing material editing.

As shown in Table \ref{tab:noise_comparison}, the degradation in image quality metrics (PSNR, SSIM, LPIPS) when introducing 50\% noise is minimal. 
For example, PSNR drops only by $0.29$ dB. 

\begin{table}[htbp]
    \centering
    \captionsetup{width=\linewidth} 
    \caption{Quantitative robustness analysis. Comparison between GT(w/ full geometry) and edited results w/ and w/o noisy thickness ($\pm 50\%$). High similarity metrics confirm perceptual robustness.}
    \label{tab:noise_comparison}
    \begin{tabular}{lccc}
    \hline
    Condition & PSNR $\uparrow$ & SSIM $\uparrow$ & LPIPS $\downarrow$ \\
    \hline
    w/o noise & 22.23 & 0.852 & 0.226 \\
    w/ noise & 21.94 & 0.849 & 0.222 \\
    \hline
    \end{tabular}
\end{table}

\section{Material Editing Settings} \label{sec:opt_set}
The selected images are out-of-domain for the indoor dataset, resulting in MatNet predictions where only albedo is relatively accurate, while roughness and metallic predictions are poor and unsuitable as initial conditions for optimization. Therefore, specific optimization conditions must be set for each image to achieve satisfactory results.

\textbf{Cherry.} The predicted roughness and metallic are relatively inaccurate, leading to excessive red light in the envmap if directly used for optimization. To address this, we initialize the envmap as pure white and start with material property optimization. The default initial values for roughness and metallic are set to 0.5. After optimizing roughness and metallic, we proceed to envmap optimization, followed by albedo optimization, as albedo predictions are more reliable.

\textbf{Bottle.} MatNet predictions are relatively accurate, and optimization follows the method described in the main paper.

\textbf{Charizard and Vase.} Similar to Cherry, we initialize the envmap as pure white and start with material property optimization. Roughness and metallic are initialized to 0.5, while albedo uses MatNet predictions. Optimization begins with roughness and metallic, followed by envmap optimization, and finally albedo optimization. During editing, setting all mask values uniformly for metallic without adjusting roughness can introduce artifacts. To improve material editing results, we use a mask during optimization. Specifically, SAM2 is used to segment Charizard’s and vase's mask, roughness and metallic values are unified within the mask during optimization. This ensures better editing results.

\end{appendices}

\end{document}